\documentclass[10pt,twocolumn,letterpaper]{article}

\usepackage{cvpr}              

\usepackage{graphicx}
\usepackage{amsmath}
\usepackage{amssymb}
\usepackage{booktabs}

\usepackage[ruled]{algorithm2e}
\usepackage{bbm}
\usepackage{bm}
\usepackage{color}
\usepackage{multirow}
\usepackage{array}
\usepackage{booktabs}
\usepackage{diagbox}
\usepackage[table]{xcolor}
\usepackage{bbding}

\usepackage[pagebackref,breaklinks,colorlinks]{hyperref}
\usepackage[accsupp]{axessibility}  

\usepackage[capitalize]{cleveref}
\crefname{section}{Sec.}{Secs.}
\Crefname{section}{Section}{Sections}
\Crefname{table}{Table}{Tables}
\crefname{table}{Tab.}{Tabs.}

\newcommand{\Tref}[1]{Table~\ref{#1}}

\newcommand{\Fref}[1]{Figure~\ref{#1}}
\newcommand{\Sref}[1]{Section~\ref{#1}}

\newcommand{\eref}[1]{Eq.~(\ref{#1})}

\newcommand{\sref}[1]{Sec.~\ref{#1}}

\def\cvprPaperID{1070} 
\def\confName{CVPR}
\def\confYear{2023}

\begin{document}

\title{Diversity-Aware Meta Visual Prompting}

\author{First Author\\
Institution1\\
Institution1 address\\
{\tt\small firstauthor@i1.org}
\and
Second Author\\
Institution2\\
First line of institution2 address\\
{\tt\small secondauthor@i2.org}
}

\author{
Qidong Huang\textsuperscript{\rm 1} \quad
Xiaoyi Dong\textsuperscript{\rm 1} \quad
Dongdong Chen\textsuperscript{\rm 2} \quad
Weiming Zhang\textsuperscript{\rm 1,}\thanks{Corresponding author.} \quad \\
Feifei Wang\textsuperscript{\rm 1} \quad 
Gang Hua\textsuperscript{\rm 3} \quad
Nenghai Yu\textsuperscript{\rm 1} \\
\textsuperscript{\rm 1}University of Science and Technology of China\quad \
\textsuperscript{\rm 2}Microsoft Cloud AI\quad \ 
\textsuperscript{\rm 3}Wormpex AI Research \ \\
{\tt\small\{hqd0037@mail., dlight@mail., zhangwm@, wangfeifei@mail., ynh@\}ustc.edu.cn} \\
{\tt\small\{cddlyf, ganghua\}@gmail.com}
}

\maketitle

\begin{abstract}
We present Diversity-Aware Meta Visual Prompting~(DAM-VP), an efficient and effective prompting method for transferring pre-trained models to downstream tasks with frozen backbone. 
A challenging issue in visual prompting is that image datasets sometimes have a large data diversity whereas a per-dataset generic prompt can hardly handle the complex distribution shift toward the original pretraining data distribution properly.
To address this issue, we propose a dataset \textbf{D}iversity-\textbf{A}ware prompting strategy whose initialization is realized by a \textbf{M}eta-prompt. Specifically, we cluster the downstream dataset into small homogeneity subsets in a diversity-adaptive way, with each subset has its own prompt optimized separately.  
Such a divide-and-conquer design reduces the optimization difficulty greatly and significantly boosts the prompting performance. 
Furthermore, all the prompts are initialized with a meta-prompt, which is learned across several datasets. It is a bootstrapped paradigm, with the key observation that the prompting knowledge learned from previous datasets could help the prompt to converge faster and perform better on a new dataset.
During inference, we dynamically select a proper prompt for each input, based on the feature distance between the input and each subset.
Through extensive experiments, our DAM-VP demonstrates superior efficiency and effectiveness, clearly surpassing previous prompting methods in a series of downstream datasets for different pretraining models. 
Our code is available at: \url{https://github.com/shikiw/DAM-VP}.

\end{abstract}

\section{Introduction}
\label{sec:intro}
With the increasing scale of training data and model size, the pretraining-finetuning paradigm has shown remarkable achievement in many areas, including natural language processing (NLP)~\cite{devlin2018bert,brown2020language} and computer vision (CV)\cite{Chen21mocov3,chen2020simclr,bao2021beit,he2022masked}. However, fully finetuning a large pre-trained model for each small downstream task still has some problems in real-world usage. The most practical one is the storage and distribution problem that we have to maintain an independent copy of the model for each task, which is quite expensive and inflexible, especially for increasing numbers of downstream tasks\cite{chen2021empirical}.


To break the dilemma, many efforts\cite{guo2021parameter,hu2021lora,he2021towards,zaken2022bitfit,chen2022adaptformer} have been paid to efficiently transfer the given pre-trained models into a particular dataset. Prompting is an extensively studied method in the NLP area, which appends a few tokens before the input sequence to provide some task-specific knowledge to the pre-trained model, so that the model could adapt well on the downstream tasks without the fully-finetuning. 
Inspired by the success of prompting in NLP, some recent works~\cite{bahng2022vp,jia2022vpt} propose visual prompting for vision models. By adding some learnable noise onto the input image or appending some learnable tokens to the model input sequence, the pre-trained models show promising results on different kinds of downstream tasks.

\begin{figure}
\centering
\includegraphics[width=0.98\linewidth]
{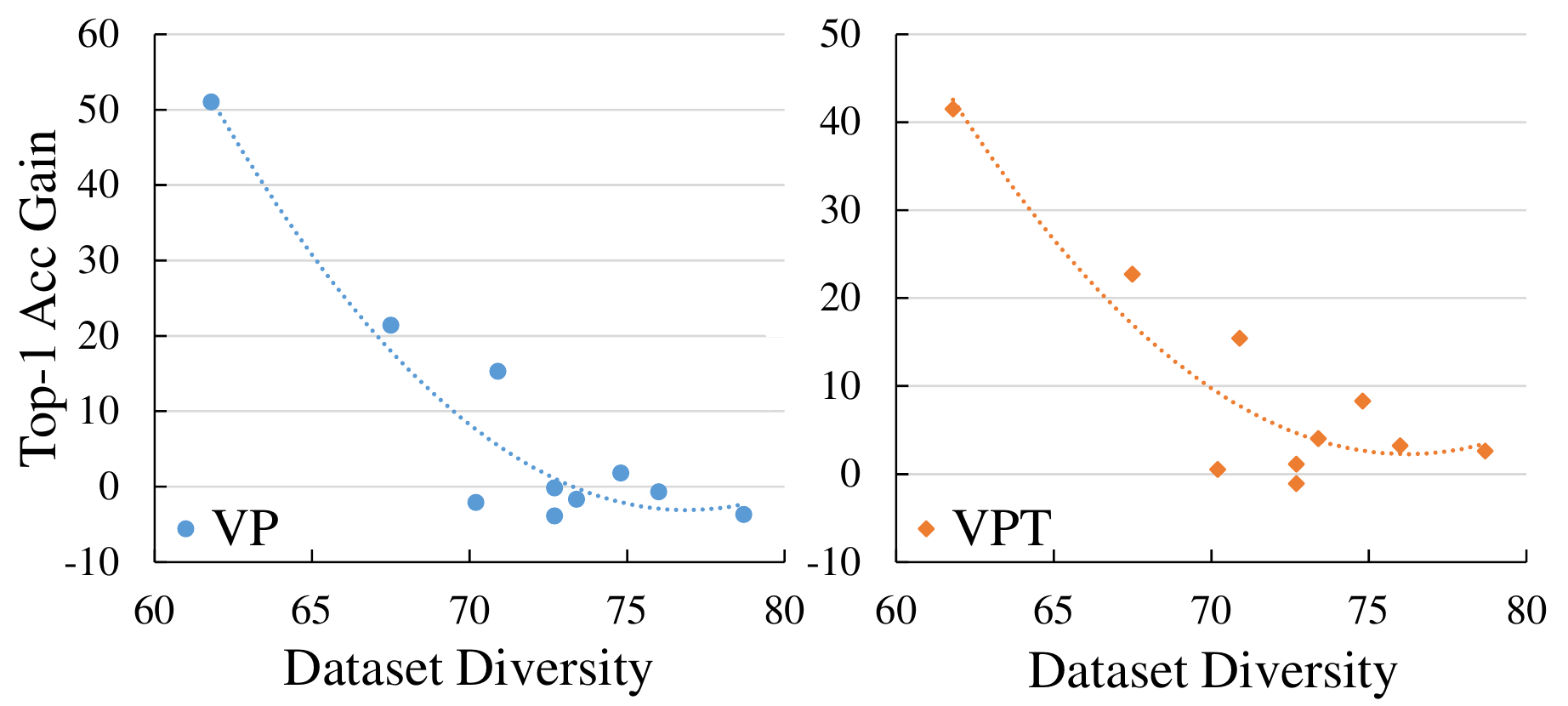}

\captionof{figure}{Relation between dataset diversity and the performance gain got by using prompting. The gain is the performance improvement when compared with the linear-probing accuracy, under the head-tuning setting. Both previous methods get a large performance gain on low-diversity datasets, while failing to boost the transfer performance on high-diversity datasets.}

\label{fig:vis}
\end{figure}

However, we argue that these methods ignore the diverse distribution property of the image dataset and using a single prompt for all the images in each dataset is not optimal. In \Fref{fig:vis}, we show the relationship between the gain from prompting and the diversity of the dataset. Here the gain represents the accuracy improvement  compared with the linear probing setting.
We find that both VP \cite{bahng2022vp} and VPT\cite{jia2022vpt} improve the model accuracy by a large margin on the low-diversity dataset, but relatively small gains on the high-diversity datasets, which is intuitively sensible. For low-diversity datasets, such as the street view house number dataset (SVHN) \cite{svhn}, all the images have similar content so a unified prompt is sufficient. On the contrary, when it comes to high-diversity datasets, such as the ImageNet~\cite{deng2009imagenet} dataset, it covers very diverse classes from the wordnet and there is not any pre-defined relationship between the classes, so it is hard to use a single prompt to provide the prior for all the images, such as for ``car'' and ``dog''.

Motivated by this observation, we propose our Diversity-Aware Meta Visual Prompting (DAM-VP). 
It has two core designs. Firstly, to provide a proper prompt for each image from high-diversity datasets, we propose a clustering-based prompt selection method. In detail, given a pre-trained visual model and a downstream dataset, we use the off-the-shelf clustering method to cluster the feature of the downstream data into several coarse-grained subsets, and guide each cluster to learn its own prompt separately. 
Based on the strong homogeneity of the same clustered data, the optimization of cluster-specific visual prompts can be greatly facilitated and the data commonalities can be also easily covered.
Secondly, we argue the prompt across different clusters or datasets may have some shared pattern, from which the model can be adapted to a new dataset faster and get better performance. This motivates us to introduce a meta-learning-based method that learns a meta prompt and initializes the prompt of each cluster with it. 

We conduct our experiments on datasets with different data diversity and evaluate the transfer performance with different pre-trained models. We report the performance on both the widely used head-tuning setting and a more challenging head-freezing/missing setting. Our DAM-VP outperforms previous methods by a large margin, especially on high-diversity datasets. For example, with the ImageNet-22k pre-trained ViT-B model, DAM-VP gets $73.1\%$ top-$1$ accuracy under the head-tuning setting on the diverse DTD~\cite{dtd} dataset, surpassing previous methods VP~\cite{bahng2022vp} and VPT~\cite{jia2022vpt} with $+13.6\%$ and $+7.3\%$ respectively.
Meanwhile, we find DAM-VP is quite efficient that with only 10 epoch tuning, it gets $85.7\%$ average top-$1$ accuracy over the 10 datasets, comparable with previous methods that tunes 100 epochs ($83.4\%$ for VP~\cite{bahng2022vp} and $85.5\%$ for VPT~\cite{jia2022vpt}).
Our contributions can be summarized as follows:
\begin{itemize}
    \item We analyze the limitation of previous visual prompting methods, and point out that vision-suitable prompting should consider the dataset diversity.
    \item Accordingly, we propose a novel Diversity-Aware Meta Visual Prompting~(DAM-VP) method. It uses the divide-and-conquer idea by clustering high-diversity datasets into subsets and learning separate prompts for each subset, in cooperation with a  meta-prompt learning design.
    \item Through extensive experiments, our DAM-VP demonstrates superior performance, achieving SOTA performance in a series of downstream datasets for different pretraining models.
\end{itemize}
 

\section{Related Work}

\noindent\textbf{Prompt learning.} 
Served as a new paradigm, prompting \cite{liu2021pre} originally emerges in NLP for adapting pre-trained language models (PLM) \cite{devlin2018bert,brown2020language} to downstream tasks. 
Its principle idea is to reformulate downstream data into the model knowledge learned during the pretraining phase, enabling the frozen pre-trained model to understand the task rather than tuning the model parameters for adaption. 
This goal has been initially reached through constructing pure text prompts that contains task-specific templates and label words to perform cloze test, \eg, hand-craft prompts \cite{gao2021making} and generative text prompts \cite{jiang2020can,shin2020autoprompt}, but unfortunately, still requiring specialized linguistic expertise for preparation. 
To alleviate this, recent efforts have been paid on prompt tuning (PT) \cite{li2021prefix,lester2021power} that learns a task-specific continuous vector as tunable prefix tokens. 
These tokens can be optimized via gradients to act as prompts in task adaption while maintaining the pre-trained model untouched. 
Driven by the success of language prompts, a lot of works \cite{radford2021learning,lu2022prompt,yao2021cpt,sung2022vl,ni2022expanding,zhang2022tip}, like CoOP \cite{zhou2022learning} and CoCoOP \cite{zhou2022conditional}, have been mushroomed to explore vision-related prompting especially in multi-modal scenarios, while still concentrating on text prompting in practice. 
Due to the gap of information density \cite{he2022masked} between languages and images, prompting for vision models is more challenging and complex. 
Inspired by prefix tuning, VPT \cite{jia2022vpt} takes the first step to adapt vision transformers to downstream tasks by prepending a set of learnable tokens at the model input. 
Concurrently, VP \cite{bahng2022vp} follows the pixel-level perspective to optimize task-specific patches that are incorporated with input images. 
Despite the pioneering successes of VP and VPT, we find that they need to pre-assign the number of prompts, which is not flexible to handle the datasets with different diversities. 
In contrast, our method uses a diversity-adaptive solution to well address this issue.

\begin{figure*}[!h]
\centering
\includegraphics[width=1.0\linewidth]{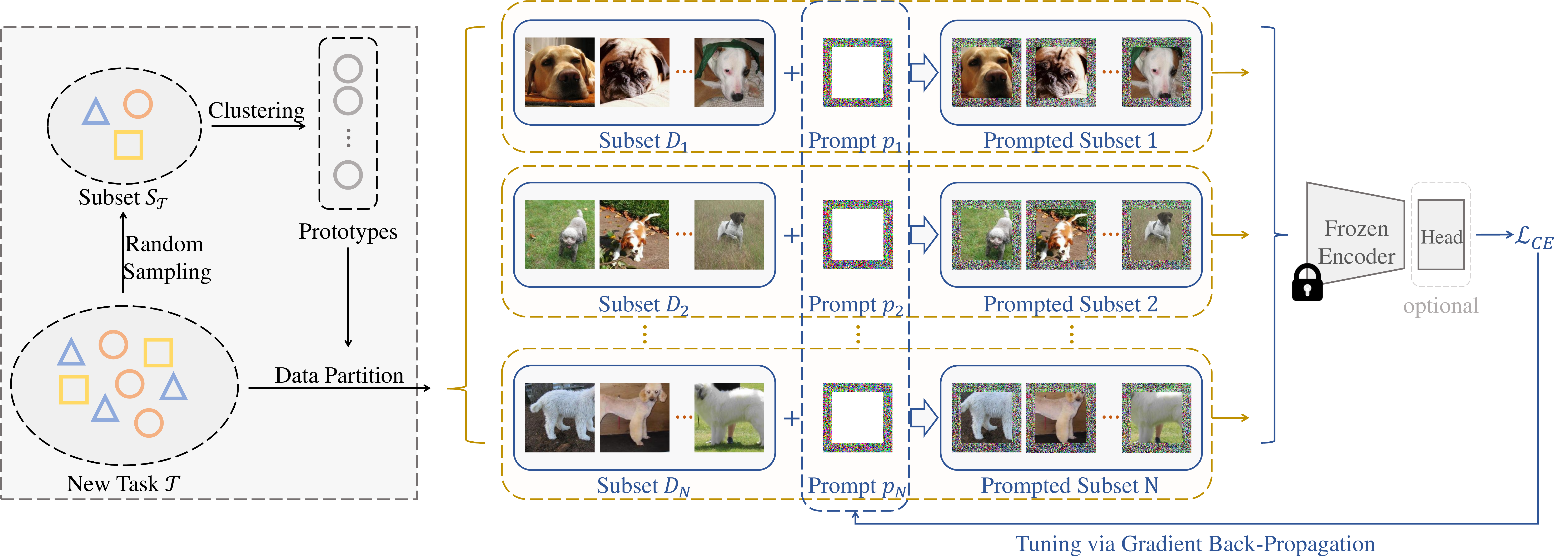}
\vspace{-2em}
\caption{The pipeline of our diversity-aware adaption for frozen pre-trained encoder $\mathcal{M}$ on new task $\mathcal{T}$. We randomly select a little subset $\mathcal{S}_\mathcal{T}$ from the training data of task $\mathcal{T}$ to implement agglomerative clustering and simulate prototypes of clusters. These prototypes are utilized to partition the whole training set into different subsets, so that we can optimize the prompt for each subset separately.}
\label{fig:pipeline_da}
\end{figure*}

\noindent\textbf{Transfer learning.} 
Typically, transfer learning focuses on how to efficiently fine-tune the supervised or un-/self-supervised pre-trained model when tackling with a new task. 
A conventional art of transfer learning is to fully fine-tune all of the model parameters on training data of the new task, using the pretraining knowledge as model initialization. 
However, the growing model capacity has exposed the inefficiency of fully fine-tuning, simulating the demand of parameter-efficient learning on downstream tasks, \ie, selecting or appending a few parameters for tuning and freezing the remaining of the model meanwhile. 
This topic has been widely explored in NLP by prepending extra learnable tokens or feature vectors \cite{li2021prefix,lester2021power} at transformer input, whereas limited in vision which still focuses on ConvNets \cite{zhang2020side,cai2020tinytl,rebuffi2017learning} and rarely on the emerging vision transformers. 
To mitigate this gap, recent efforts have explored to efficiently transfer vision transformers by introducing a parallel trainable down-to-up branch into the MLP module of self-attention blocks \cite{chen2022adaptformer} or scaling and shifting the learned features \cite{lian2022scaling}. 
VPT \cite{jia2022vpt} is the pioneer work to leverage learnable prefix tokens/features for visual prompting, but still not efficient in terms of convergence time. 
Driven by this, our DAM-VP strives to learn faster during tuning with comparable amount of learnable parameters introduced.


\section{Method}
\label{sec:method}

\begin{figure*}[h]
\centering
\includegraphics[width=1.0\linewidth]{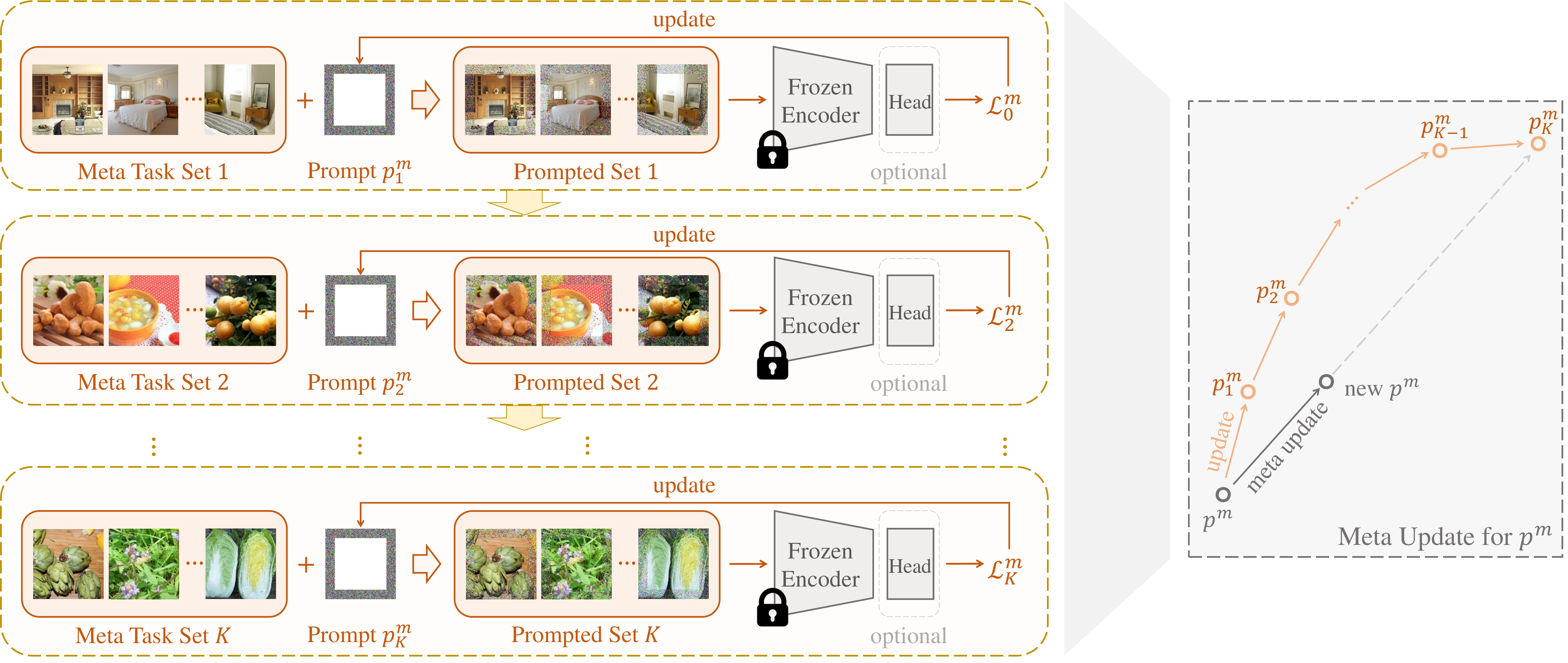}
\vspace{-1.5em}
\caption{The pipeline of meta-learning-based prompt initialization. We partition each task dataset into subsets and regard each subset as a single meta task. During meta training on each meta batch, we continuously update the temporary prompt on each meta task set and finally apply moving average to get the updated meta prompt.}
\label{fig:pipeline_m}
\end{figure*}

We introduce our Diversity-Aware Meta Visual Prompting~(DAM-VP), a novel prompting method that is adaptive to diverse downstream datasets effectively and efficiently. As shown in \Fref{fig:pipeline_da}, with a given dataset, DAM-VP first extracts its specific prototypes in an unsupervised and adaptive manner as the pre-processing. Then we split the dataset into different subsets, according to the prototypes. For each subset, we assign a specific prompt to it and optimize it with the tuning loss. Rather than random initialization, all of the prompts are initialized by a meta prompt learned across different datasets, as shown in \Fref{fig:pipeline_m}.

\begin{figure}[t]
\centering
\begin{minipage}{0.19\linewidth}
    \centering
    \includegraphics[width=1\linewidth]{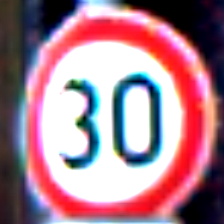}
\end{minipage}
\hfill
\begin{minipage}{0.19\linewidth}
    \centering
    \includegraphics[width=1\linewidth]{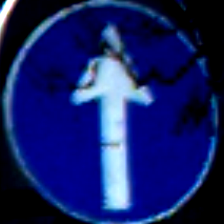}
\end{minipage}
\hfill
\begin{minipage}{0.19\linewidth}
    \centering
    \includegraphics[width=1\linewidth]{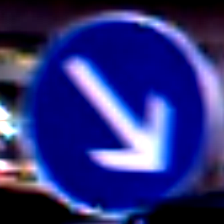}
\end{minipage}
\hfill
\begin{minipage}{0.19\linewidth}
    \centering
    \includegraphics[width=1\linewidth]{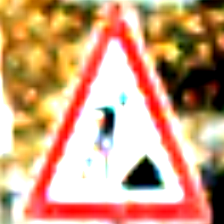}
\end{minipage}
\hfill
\begin{minipage}{0.19\linewidth}
    \centering
    \includegraphics[width=1\linewidth]{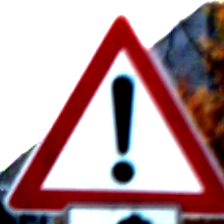}
\end{minipage}
\vfill
\begin{minipage}{0.19\linewidth}
    \centering
    \includegraphics[width=1\linewidth]{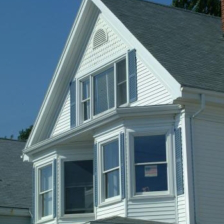}
\end{minipage}
\hfill
\begin{minipage}{0.19\linewidth}
    \centering
    \includegraphics[width=1\linewidth]{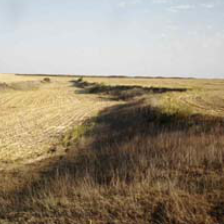}
\end{minipage}
\hfill
\begin{minipage}{0.19\linewidth}
    \centering
    \includegraphics[width=1\linewidth]{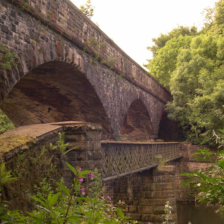}
\end{minipage}
\hfill
\begin{minipage}{0.19\linewidth}
    \centering
    \includegraphics[width=1\linewidth]{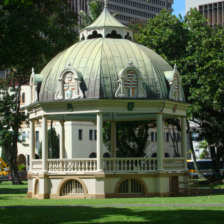}
\end{minipage}
\hfill
\begin{minipage}{0.19\linewidth}
    \centering
    \includegraphics[width=1\linewidth]{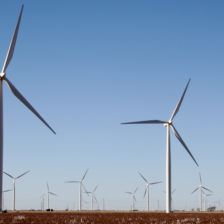}
\end{minipage}
\caption{Examples of GTSRB (\textbf{top}) and SUN397
(\textbf{bottom}).}
\label{fig:sun_vs_gtsrb}
\end{figure}

\noindent\textbf{Diversity-adaptive dataset partition.}
As we briefly introduced in \sref{sec:intro}, different image datasets have different distribution diversity. 
Take \Fref{fig:sun_vs_gtsrb} as an example, when comparing with the traffic sign dataset\cite{gtsrb}, the data in the scene dataset \cite{sun397} is more diverse in terms of angle, illumination, the complexity of content, \etc. 
The prompting is designed to reduce the distribution gap between the target downstream dataset and the model pretraining data, and it is intuitive that a dataset with similar content is easy to transfer. So it is a straightforward idea to split a diverse dataset into small subsets and apply different prompts to each subset.

To this end, we consider the diversity property in our visual prompting design and propose an adaptive dataset partition strategy to suit the diversity of task data automatically. 
Specifically, when adapting the frozen pre-trained backbone $\mathcal{M}$ to the new task $\mathcal{T}$, we first randomly sample a small subset $\mathcal{S}_\mathcal{T}$ from the training set of $\mathcal{T}$. 
Then we extract the feature of subset $\mathcal{S}_\mathcal{T}$ with the frozen $\mathcal{M}$ without any prompting. We denote the features as $\{\mathcal{M}(s) | s\in \mathcal{S}_\mathcal{T}\}$ and use an off-the-shelf clustering method\cite{mullner2011modern} to aggregate them into several clusters. 
The clustering procedure is time-efficient (less than 1\% of total tuning time) and the number of clusters $N$ is auto-adaptive to the dataset diversity by a pre-defined threshold. 
Once the clusters are constructed, we compute the average values for features of each cluster as the cluster-specific prototypes $\{c_i\}_{i=1}^N$, \ie, 
\begin{equation}
    c_i = \frac{1}{|\mathcal{S}_i|}\sum_{s\in \mathcal{S}_i} \mathcal{M}(s), \quad i = 1, \cdots, N, 
\end{equation}
where $\mathcal{S}_i$ is the data samples corresponding to the $i^{th}$ cluster that satisfies $|\mathcal{S}_1| + |\mathcal{S}_2| + \dots + |\mathcal{S}_N| = |\mathcal{S}_\mathcal{T}|$. 
The unsupervised mechanism naturally guarantees that the dataset with a higher diversity can be divided into more clusters, and the dataset with a lower diversity can be divided into fewer clusters or even just a single cluster. 
In practice, we configure the threshold of clustering properly to keep $N$ in a reasonable range, usually less than the data category number.

\noindent\textbf{Diversity-aware prompt selection.}
With the simulated prototypes $\{c_i\}_{i=1}^N$, it is easy to partition the whole training dataset $\mathcal{D}_\mathcal{T}$ of task $\mathcal{T}$ into small subsets $\{\mathcal{D}_i\}_{i=1}^N$. 
Correspondingly, we assign one visual prompt for each subset and get $N$ visual prompts $\{p_i\}_{i=1}^N$ in total.
Similar to the design in ~\cite{bahng2022vp}, we use a photo-frame-like pixel-level prompting. It has the same size as the model input and we add it to the input image directly. Such a design has two advantages: Firstly, it does not introduce additional computation cost during inference, while the prefix-token prompting used in VPT \cite{jia2022vpt} increases the input length and leads to a larger computation cost.  Secondly, such pixel-level prompting is irrelevant to the model type, so it could be used in both recent popular Vision Transformer models and traditional convolution networks.  On the contrary, the prefix-token prompting is specifically designed for token-list-type input so that could only be used for Vision Transformer.

Given an image-label pair $(x,y)$ from the training set, we forward $x$ on the frozen $\mathcal{M}$ without prompting to get its original feature and compute the euclidean distances between the feature and each aforementioned prototype. 
The prompt that corresponds to the minimal distance is considered as the prompt added on $x$. Formally, prompted image $x^p$ is defined as
\begin{equation}
    x^p \triangleq x + p_t, \quad \text{s.t.}\quad t = \mathop{\arg\min}_i \| \mathcal{M}(x) - c_i \|_2^2.
\end{equation}
which indicates the input image $x$ is assigned into the $t^{th}$ image subset $\mathcal{D}_t$.

\noindent\textbf{Prompt learning.}
In this paper, we seek a more general prompting format that could be utilized in different settings. 1) The head-tuning setting used in VPT~\cite{jia2022vpt} that a learnable classification head is optimized with the prompt jointly. 2) The head-freezing/missing setting that only the prompt is learnable. The second one is a more challenging task, but it is also a more flexible real-world usage format: we only need to add different prompts on the input for different tasks, maintaining an end-to-end frozen pre-trained model.

For the head-tuning setting, we follow the design in VPT~\cite{jia2022vpt} that optimize a new $k$-class classification head for the target task with $k$ categories. 
Similarly, for the head-freezing case, we assign the first $k$ classes in the frozen head to the new task.
When it comes to the head-missing case that the pre-trained model does not have a classification head (such as self-supervised pre-trained models), a simple hard-coded mapping solution~\cite{bahng2022vp} is convert the output feature (\eg, 768 channels output feature of ViT/16-Base \cite{DosovitskiyB0WZ21} encoder) to classification logits. However, we argue such hard-coded mapping is inefficient, because the optimization might be limited if the neurons at these fixed positions are not active enough. So we propose an active-based mapping method, which selects the most active top-$k$ neurons of the output layer of $\mathcal{M}$ by measuring the variances of each position output when confronted with random noise inputs.

Based on the above designs, we can minimize the cross-entropy loss between the logits of prompted image $x^p$ and groud-truth label $y$ to tune our visual prompts on $\mathcal{D}_{\mathcal{T}}$, \ie,
\begin{equation}
    p_1^\ast, \dots, p_N^\ast = \mathop{\arg\min}_{p_1, \dots, p_N} \frac{1}{|\mathcal{D}_\mathcal{T}|} \sum_{i=1}^N \sum_{x\in \mathcal{D}_i} \mathcal{L}_{CE} (\mathcal{M}(x + p_i), y). 
\end{equation}

Once the prompts are well-optimized, we can utilize prototypes to categorize the input image and assign it to its belonging subset during inference. 
The corresponding prompt is subsequently incorporated with the input image as prompted input to get the classification logits. 

\noindent\textbf{Prompt boosting via Meta-learning.}
In real-world usage, a frozen model $\mathcal{M}$ needs to be transferred to a bunch of downstream tasks, this leads to two desired favorable properties of the prompting method. First, it should be efficient that only a few epochs tuning could get a good result. Second, the prompts learned from previous tasks could help the new tasks learn better prompts so that the method is bootstrapped.

In light of meta-learning \cite{finn2017model,nichol2018reptile}, we integrate a quick algorithm Reptile \cite{nichol2018reptile} into our diversity-aware learning to boost the prompting learning. 
The principle idea is to learn a meta prompt $p^m$ on several task datasets $\{\mathcal{D}_i^m\}_{i=1}^M$ that are prepared as the meta training data, and adopt the well-trained $p^m$ as the initial prompt for diversity-aware adaption. 
Specifically, we first partition each task dataset into subsets, \eg, dataset $\mathcal{D}_i^m$ is divided into $\mathcal{D}_{i,1}^m, \mathcal{D}_{i,2}^m, \dots, \mathcal{D}_{i,K_i}^m$. 
Then, we regard each subset as a single meta task and sample images from each subset to form the meta training batch $\mathcal{B}$, \ie, dataset $\mathcal{D}_i^m$ contributes total $K_i$ meta task sets for $\mathcal{B}$. 
Formally, a meta training batch is constructed by
\begin{equation}
    \mathcal{B} = \bigcup_{j=1}^K \mathcal{B}_j \quad \text{s.t.}\quad \mathcal{B}_j \in \mathcal{G}_j.
\end{equation}
where $K$ is the total number of subsets that satisfies $K = K_1 + K_2 + \dots + K_M$ and we rename each subset as group $\mathcal{G}_j$ for convenience, satisfying $\bigcup_{i=1}^M \bigcup_{k=1}^{K_i} \mathcal{D}_{i,k}^m = \bigcup_{j=1}^K \mathcal{G}_j$.

With the sampled meta training batch $\mathcal{B}$, we can update the temporary prompt $p_j^m$ on each meta task set $\mathcal{B}_j$ by minimizing cross-entropy loss $\mathcal{L}_j^m$, \ie,
\begin{equation}
\begin{split}
    p_j^m &= p_{j-1}^m - \eta\nabla_{p_{j-1}^m} \frac{1}{|\mathcal{B}_j|}\sum_{x\in\mathcal{B}_j}\mathcal{L}_j^m, 
    \\
    \text{s.t.}\quad \mathcal{L}_j^m &= \mathcal{L}_{CE} (\mathcal{M}(x+p_{j-1}^m), y),
\label{eq:fast_update}
\end{split}
\end{equation}
where $y$ is the ground-truth label of $x$. Finally, we implement meta update to get the new meta prompt after training on meta batch $\mathcal{B}$ by moving average, 
\begin{equation}
    p^m \gets p^m + \gamma\frac{1}{K} \sum_{j=1}^K (p_j^m - p^m),
\label{eq:meta_update}
\end{equation}
where $\gamma$ is the meta update step that varies within $(0,1)$.


\section{Experiments}

\subsection{Setup}
\label{sec:setup}

\noindent\textbf{Datasets.} 
Here we select 16 popular image datasets for the experiments including CIFAR10 \cite{cifar}, CIFAR100 \cite{cifar}, DTD \cite{dtd}, CUB200 \cite{cub200}, NABirds \cite{nabirds}, Stanford-Dogs \cite{stanford_dogs}, Oxford-Flowers \cite{oxford_flowers}, Food101 \cite{food101}, GTSRB \cite{gtsrb}, SVHN \cite{svhn}, SUN397 \cite{sun397}, STL10 \cite{stl10}, Fru92 \cite{hou2017vegfru}, Veg200 \cite{hou2017vegfru}, Oxford-IIIT Pet \cite{parkhi2012cats} and EuroSAT \cite{helber2019eurosat}, where the first 10 datasets are used for prompt evaluation and the remaining 6 are prepared for our meta prompt initialization. 
For data preprocessing, we randomly resize the input image into the size of $256\times 256$ and subsequently crop it into $224\times 224$. 
More details including the performance on VTAB-1k benchmark \cite{zhai2019large} are provided in supplementary.

\noindent\textbf{Models.} 
Our experiments involve six pre-trained vision models including ImageNet-1k~\cite{deng2009imagenet} supervised ViT-B/16~\cite{DosovitskiyB0WZ21}, supervised ResNet-50 \cite{he2016deep}, MoCo v3 \cite{Chen21mocov3} learned ViT-B/16; ImageNet-22k supervised ViT-B/16 and Swin-Base~\cite{liu2021swin}; 400m web data contrastive learning ViT-B/16 model CLIP~\cite{radford2021learning}. As we illustrated above, our DAM-VP could be used for models without classification heads and traditional convolution networks, we discuss these settings in the supplementary materials.

\noindent\textbf{Baselines.} 
We compare our method with both parameter-tuning methods and prompt-tuning methods. For parameter-tuning, we report the fully-tuning, linear probing results as baseline, and the efficient-tuning method Adapter~\cite{adapter1,adapter2}. For prompt-tuning, we compare with the VP\cite{bahng2022vp} and VPT\cite{jia2022vpt}.

\begin{table*}[t]
    \scriptsize
    \centering
    \setlength{\tabcolsep}{1.2mm}{
    \begin{tabular}{l|p{11mm}<{\centering}|p{9mm}<{\centering}p{9mm}<{\centering}p{9mm}<{\centering}p{9mm}<{\centering}p{9mm}<{\centering}p{9mm}<{\centering}p{11mm}<{\centering}p{11mm}<{\centering}p{9mm}<{\centering}p{9mm}<{\centering}|p{11mm}<{\centering}}
        \toprule
        & \multirow{1}{*}{Extra}
        & DTD
        & CUB200 
        & NABirds 
        & Dogs 
        & Flowers
        & Food101 
        & CIFAR100
        & CIFAR10 
        & GTSRB 
        & SVHN
        & Average
        \\
        & \multirow{1}{*}{Head}
        & \cite{dtd}
        & \cite{cub200}
        & \cite{nabirds}
        & \cite{stanford_dogs}
        & \cite{oxford_flowers}
        & \cite{food101}
        & \cite{cifar}
        & \cite{cifar}
        & \cite{gtsrb}
        & \cite{svhn}
        &
        \\
        \midrule
        Data diversity & - & 78.7 & 76.0 & 74.8 & 73.4 & 72.7 & 72.7 & 70.9 & 70.2 & 67.5 & 61.8 & -
        \\
        \midrule
        Fully-Tuning & \Checkmark & 70.6 & 84.7 & 72.3 & 84.6 & 98.3 & 83.0 & 87.5 & 97.4 & 96.8 & 96.9 & 87.2
        \\
        Linear & \Checkmark & 68.7 & 83.5 & 69.3 & 84.4 & 97.7 & 78.5 & 77.6 & 92.9 & 65.6 & 61.1 & 77.9
        \\
        Adapter \cite{adapter1,adapter2} & \XSolidBrush & 47.7 & 17.5 & 3.8 & 32.0 & 40.1 & 46.1 & 43.0 & 72.8 & 82.2 & 19.6  & 40.5
        \\
        VP \cite{bahng2022vp} & \XSolidBrush & 47.8 & 40.6 & 13.8 & 61.9 & 56.5 & 55.7 & 54.4 & 92.9 & 86.0 & 87.8 & 59.7
        \\
        VPT \cite{jia2022vpt} & \XSolidBrush & 27.8 & 10.9 & 1.3 & 36.7 & 9.3 & 63.0 & 31.8 & 46.1 & 84.3 & 28.2 & 33.9
        \\
        \cellcolor{gray!20}\textbf{DAM-VP (10 epochs)} & 
        \cellcolor{gray!20}\XSolidBrush &
        \cellcolor{gray!20}51.3 & \cellcolor{gray!20}43.6 & \cellcolor{gray!20}22.3 & \cellcolor{gray!20}70.8 & \cellcolor{gray!20}65.9 & \cellcolor{gray!20}61.5 & \cellcolor{gray!20}61.5 & \cellcolor{gray!20}90.5 & \cellcolor{gray!20}79.7 &  \cellcolor{gray!20}85.7 & \cellcolor{gray!20}63.3
        \\
        \cellcolor{gray!20}\textbf{DAM-VP (50 epochs)} & 
        \cellcolor{gray!20}\XSolidBrush &
        \cellcolor{gray!20}53.9 & \cellcolor{gray!20}64.6 & \cellcolor{gray!20}38.6 & \cellcolor{gray!20}75.5 & \cellcolor{gray!20}84.1 & \cellcolor{gray!20}66.6 & \cellcolor{gray!20}67.2 & \cellcolor{gray!20}92.4 & \cellcolor{gray!20}86.2 &  \cellcolor{gray!20}88.4 & \cellcolor{gray!20}71.8
        \\
        \bottomrule
    \end{tabular}
    }
    \caption{Head-freezing/missing adaption performance of different methods on ViT-B-1K, where we report image classification accuracy and all of baseline methods are trained for 50 epochs.}
    \label{tab:table1}
\end{table*}

\begin{table*}[t]
    \scriptsize
    \centering
    \setlength{\tabcolsep}{1.2mm}{
    \begin{tabular}{l|p{11mm}<{\centering}|p{9mm}<{\centering}p{9mm}<{\centering}p{9mm}<{\centering}p{9mm}<{\centering}p{9mm}<{\centering}p{9mm}<{\centering}p{11mm}<{\centering}p{11mm}<{\centering}p{9mm}<{\centering}p{9mm}<{\centering}|p{11mm}<{\centering}}
        \toprule
        & \multirow{1}{*}{Extra}
        & DTD
        & CUB200 
        & NABirds 
        & Dogs 
        & Flowers
        & Food101 
        & CIFAR100
        & CIFAR10 
        & GTSRB 
        & SVHN
        & Average
        \\
        & \multirow{1}{*}{Head}
        & \cite{dtd}
        & \cite{cub200}
        & \cite{nabirds}
        & \cite{stanford_dogs}
        & \cite{oxford_flowers}
        & \cite{food101}
        & \cite{cifar}
        & \cite{cifar}
        & \cite{gtsrb}
        & \cite{svhn}
        &
        \\
        \midrule
        Data diversity & - & 78.7 & 76.0 & 74.8 & 73.4 & 72.7 & 72.7 & 70.9 & 70.2 & 67.5 & 61.8 & -
        \\
        \midrule
        Fully-Tuning & \Checkmark & 78.6 & 81.9 & 72.6 & 80.5 & 97.3 & 91.8 & 80.9 & 96.3 & 95.3 & 95.7 & 87.1
        \\
        Linear & \Checkmark & 77.0 & 81.5 & 72.4 & 79.5 & 95.8 & 92.2 & 79.8 & 95.0 & 85.6 & 69.2 & 82.8
        \\
        TP \cite{radford2021learning} & \XSolidBrush & 41.8 & 55.5 & 44.7 & 62.5 & 65.5 & 87.7 & 64.6 & 87.5 & 40.2 & 13.7 & 56.4
        \\
        VP \cite{bahng2022vp} & \XSolidBrush & 54.3 & 56.8 & 46.3 & 63.5 & 71.7 & 86.5 & 70.7 & 93.2 & 90.5 & 90.4 & 73.4
        \\
        \cellcolor{gray!20}\textbf{DAM-VP (10 epochs)} & 
        \cellcolor{gray!20}\XSolidBrush &
        \cellcolor{gray!20}58.4 & \cellcolor{gray!20}61.1 & \cellcolor{gray!20}49.6 & \cellcolor{gray!20}68.6 & \cellcolor{gray!20}84.5 & \cellcolor{gray!20}85.0 & \cellcolor{gray!20}68.3 & \cellcolor{gray!20}92.7 & \cellcolor{gray!20}87.6 &  \cellcolor{gray!20}87.9 &  \cellcolor{gray!20}74.4
        \\
        \cellcolor{gray!20}\textbf{DAM-VP (50 epochs)} & 
        \cellcolor{gray!20}\XSolidBrush &
        \cellcolor{gray!20}63.7 & \cellcolor{gray!20}65.9 & \cellcolor{gray!20}54.8 & \cellcolor{gray!20}71.5 & \cellcolor{gray!20}87.0 & \cellcolor{gray!20}87.7 & \cellcolor{gray!20}72.2 & \cellcolor{gray!20}93.7 & \cellcolor{gray!20}92.3 &  \cellcolor{gray!20}90.4 &  \cellcolor{gray!20}77.9
        \\
        \bottomrule
    \end{tabular}
    }
    \caption{Head-freezing/missing adaption performance of different methods on CLIP-ViT-B, where we report image classification accuracy and all of baseline methods are trained for 50 epochs. ``TP'' denotes text prompting that directly adopts zero-shot classification head of CLIP. Here ``VP'' and our method also use this fixed head rather than particular mapping to get output logits.}
    \label{tab:table2}
\end{table*}

\noindent\textbf{Diversity metrics.} 
To quantitatively measure the data diversity of a given dataset, we follow \cite{bahng2022vp} to randomly sample 10,000 image pairs in each dataset and compute the LPIPS distance \cite{zhang2018unreasonable} of each pair. 
The average LPIPS is regarded to measure the perceptual diversity of the given dataset.

\noindent\textbf{Implementation details.} 
For meta learning based prompt initialization, we unify update learning rate $\eta$ (in \eref{eq:fast_update}) as 0.5 and meta step size $\gamma$ (in \eref{eq:meta_update}) as 0.5. 
Our meta prompts of different pre-trained backbones are equally trained for 200 epochs by Adam optimizer with a cosine annealing schedule. 
The step number for updating the temporary prompt on each meta task is set as 4. 
For diversity-aware adaption, we set the subset size $|\mathcal{S}_\mathcal{T}|$ as 1000 by default. 
More implementation details, such as training or clustering configurations, are provided in our supplementary.

\subsection{Comparison with Baseline Methods}
\noindent\textbf{Quantitative results.} 
We comprehensively compare our method with the baselines mentioned in \Sref{sec:setup}. 
Two scenarios are taken into our consideration:
1) \textbf{head-freezing/missing adaption}, \ie, only tuning the introduced modules like prompts without any extra task-specific head. 
2) \textbf{head-tuning adaption}, \ie, tuning the introduced modules like prompts along with learning a task-specific head. 
\Tref{tab:table1} and \ref{tab:table2} show the quantitative comparison results of the head-freezing/missing setting between our DAM-VP and other adaption methods. 
We can find that our method significantly outperforms baseline methods even when our prompts are just trained for 10 epochs. 
\Tref{tab:table3} and \ref{tab:table4} show the quantitative comparison results of the head-tuning setting between our DAM-VP and other adaption methods. 
It is easy to discover that DAM-VP presents its strong ability to help pre-trained models generalize on various image datasets, surpassing other methods with fewer training epochs and higher recognition accuracy. Furthermore, our DAM-VP even outperforms the fully-tuning setting on both the ViT-B-22K model and the Swin-B-22K model with only 50 epochs tuning.

\begin{table}[t]
    \scriptsize
    \centering
    \setlength{\tabcolsep}{1mm}{
    \begin{tabular}{lccp{11mm}<{\centering}p{11mm}<{\centering}p{11mm}<{\centering}p{11mm}<{\centering}}
        \toprule
        \multirow{2}{*}{Setting}
        & \multirow{2}{*}{Meta}
        & \multirow{2}{*}{Diversity}
        & CUB200 
        & Flowers
        & CIFAR100
        & SVHN
        \\
        & &
        & \cite{cub200}
        & \cite{oxford_flowers}
        & \cite{cifar}
        & \cite{svhn}
        \\
        \midrule
        $\mathbf{A}$ & \XSolidBrush & \XSolidBrush & 41.2 & 59.9 & 55.1 & 87.9
        \\
        $\mathbf{B}$ & \Checkmark & \XSolidBrush & 43.9 & 65.7 & 59.4 & 88.1
        \\
        $\mathbf{C}$ & \XSolidBrush & \Checkmark & 63.3 & 81.5 & 66.9 & 88.2
        \\
        \cellcolor{gray!20}$\mathbf{D}$ & \cellcolor{gray!20}\Checkmark & \cellcolor{gray!20}\Checkmark & 
        \cellcolor{gray!20}64.6 &
        \cellcolor{gray!20}84.1 & \cellcolor{gray!20}67.2 &  \cellcolor{gray!20}88.4
        \\
        \bottomrule
    \end{tabular}
    }
    \caption{Ablation results of diversity-aware strategy and meta-prompt initialization. We report Top-1 accuracy on ViT-B-1K.}
    \label{tab:abl_1}
\end{table}

\begin{figure*}[t]
\centering
\vspace{-1.2em}
\begin{minipage}{0.245\linewidth}
    \centering
    \includegraphics[width=1\linewidth]{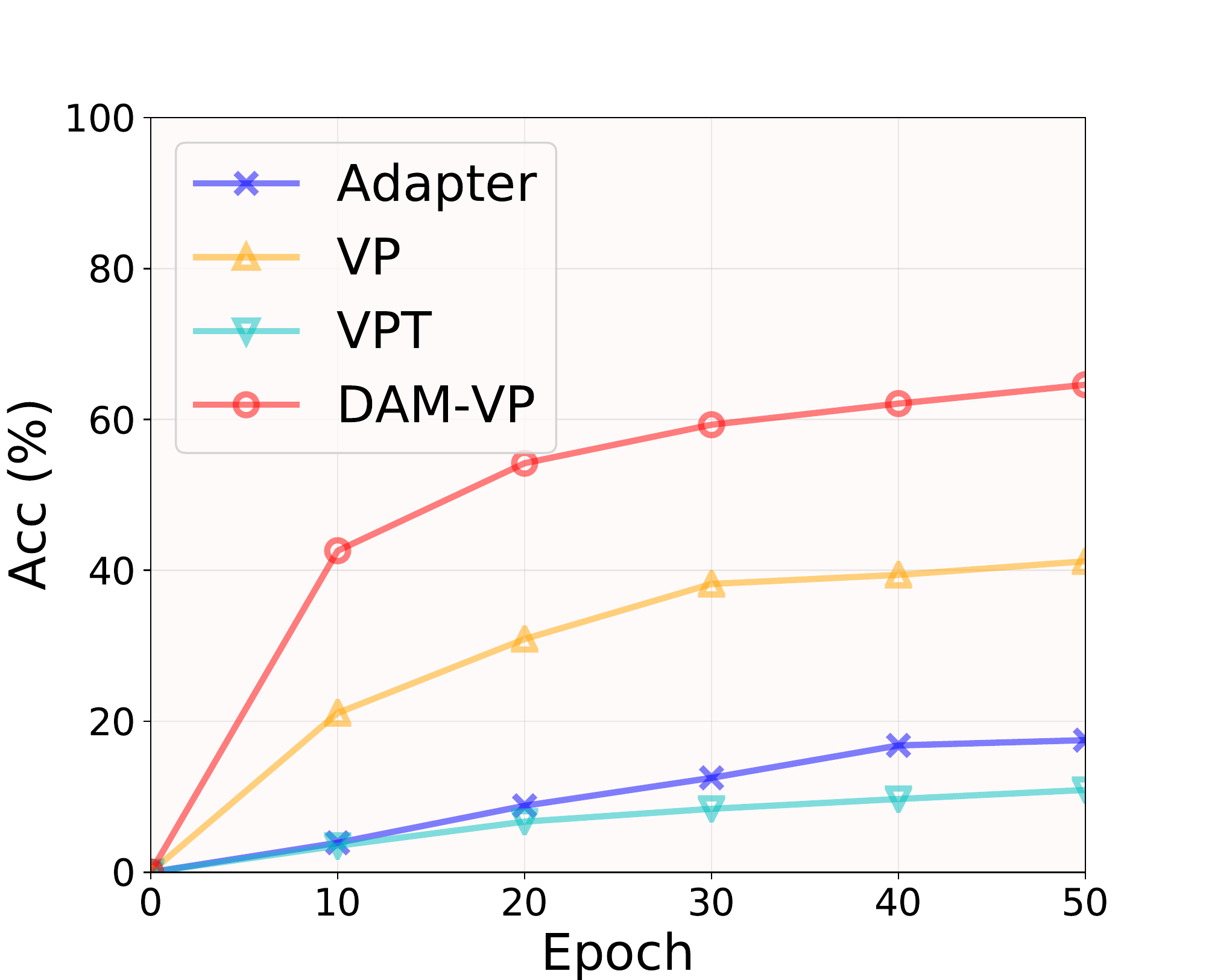}
    \\
    \footnotesize CUB200
\end{minipage}
\hfill
\begin{minipage}{0.245\linewidth}
    \centering
    \includegraphics[width=1\linewidth]{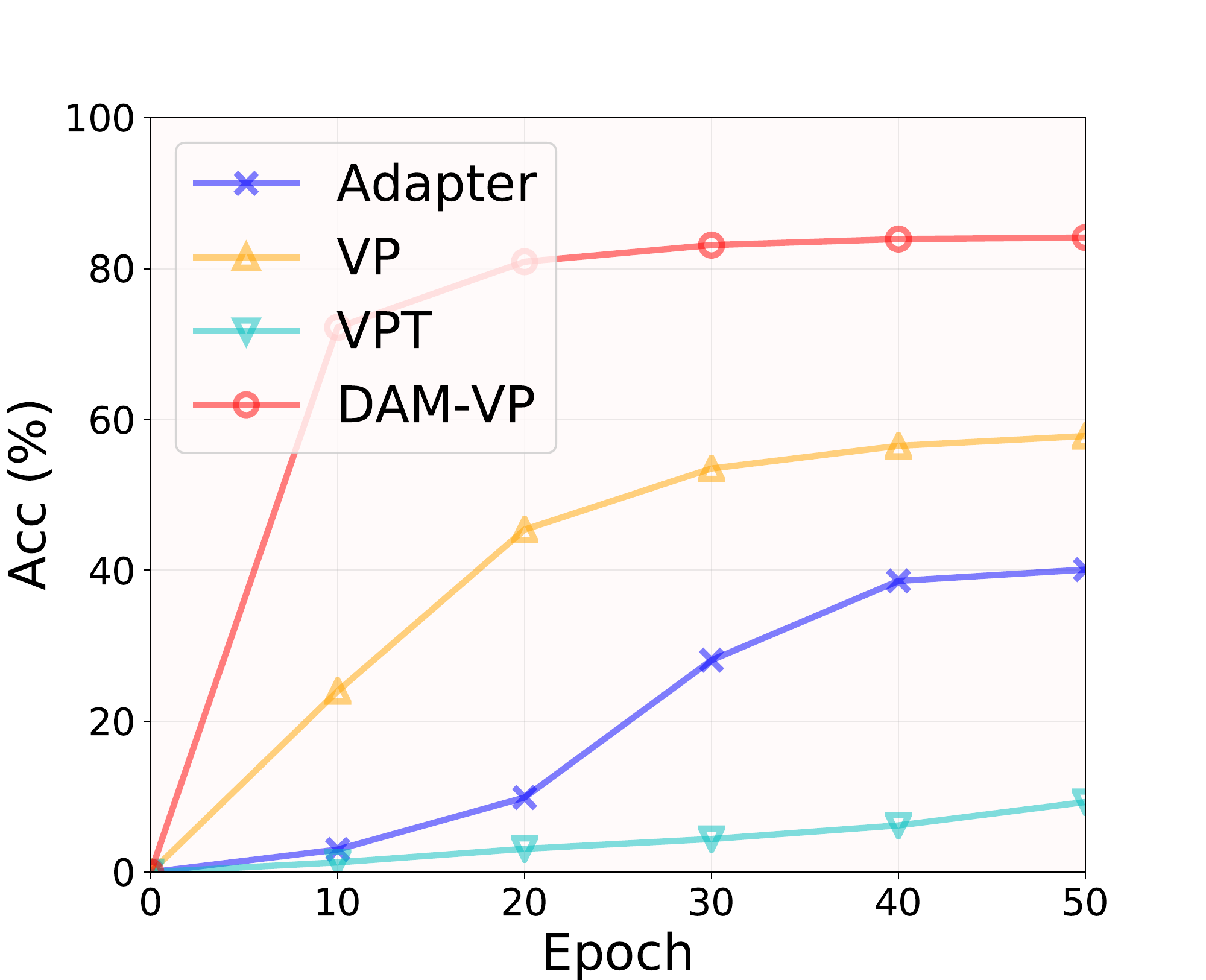}
    \\
    \footnotesize Oxford-Flowers
\end{minipage}
\hfill
\begin{minipage}{0.245\linewidth}
    \centering
    \includegraphics[width=1\linewidth]{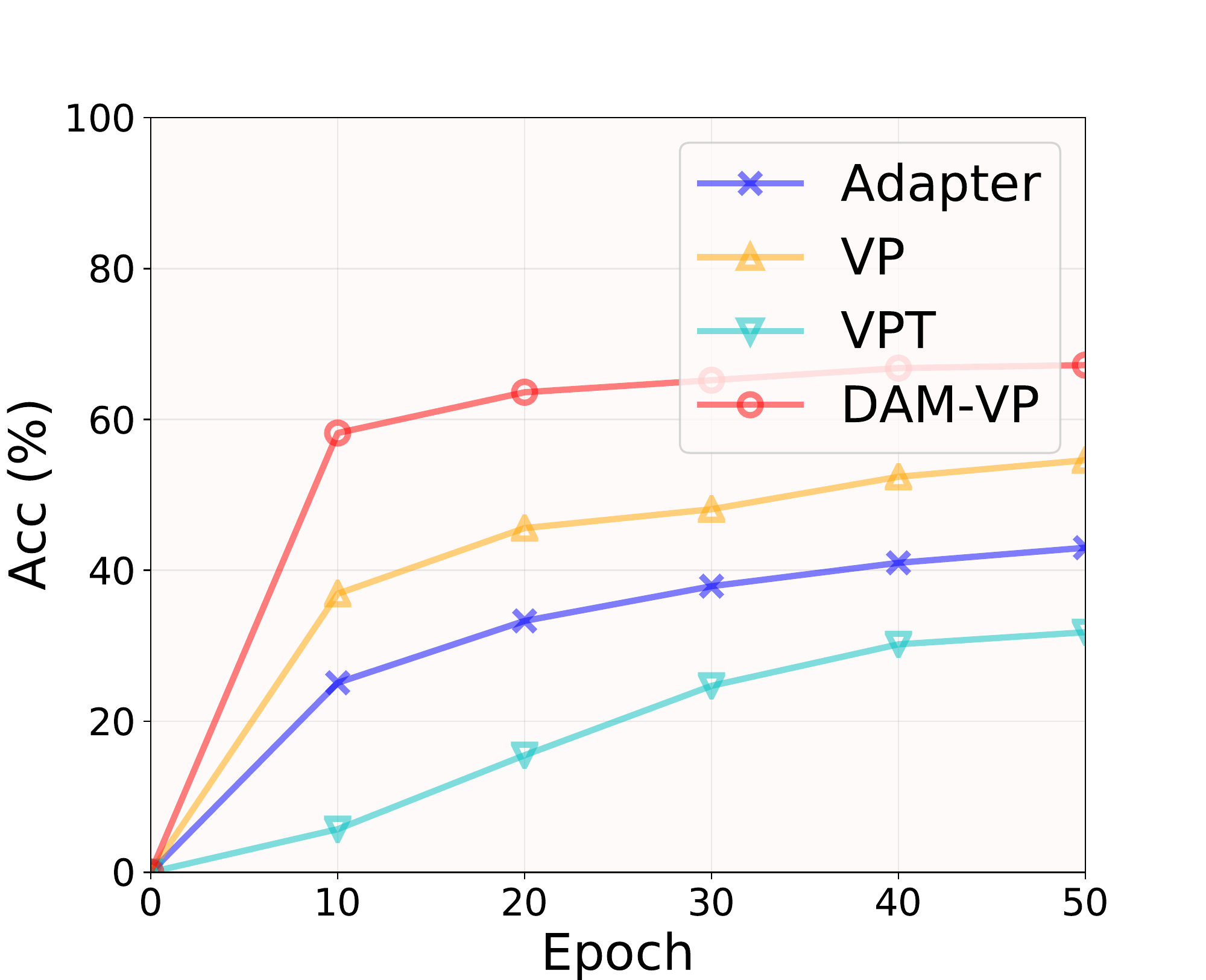}
    \\
    \footnotesize CIFAR100
\end{minipage}
\hfill
\begin{minipage}{0.245\linewidth}
    \centering
    \includegraphics[width=1\linewidth]{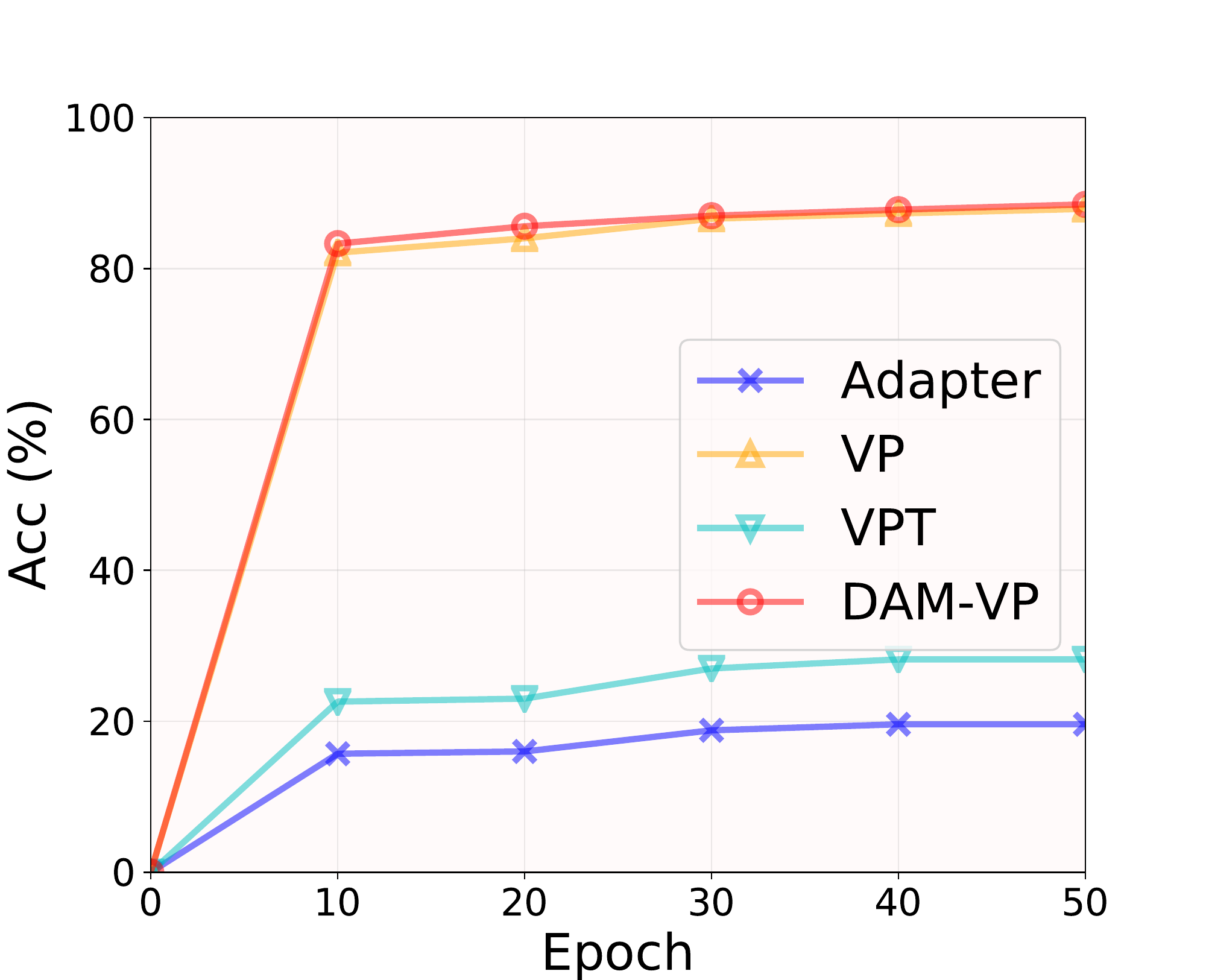}
    \\
    \footnotesize SVHN
\end{minipage}
\vfill
\begin{minipage}{0.245\linewidth}
    \centering
    \includegraphics[width=1\linewidth]{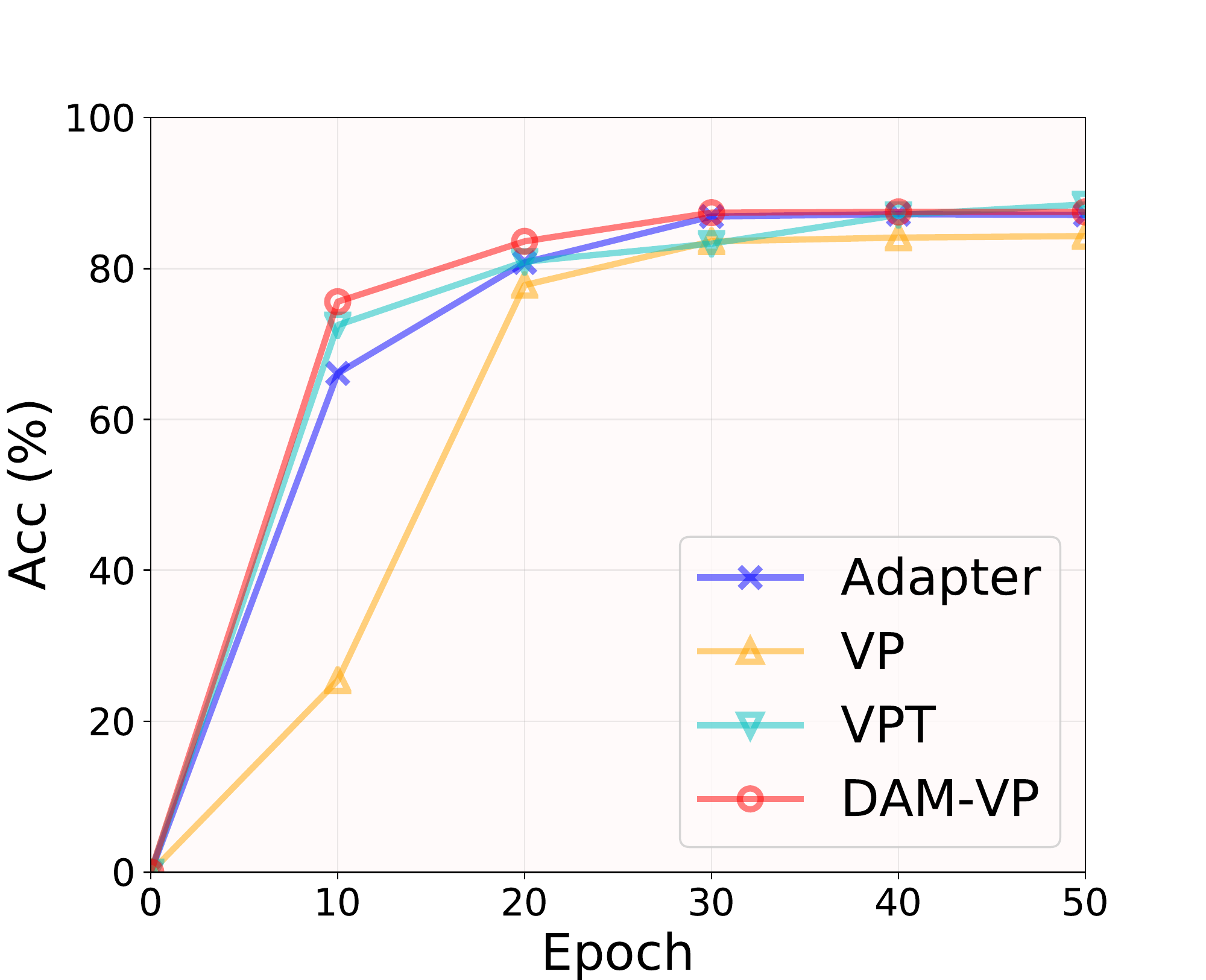}
    \\
    \footnotesize CUB200
\end{minipage}
\hfill
\begin{minipage}{0.245\linewidth}
    \centering
    \includegraphics[width=1\linewidth]{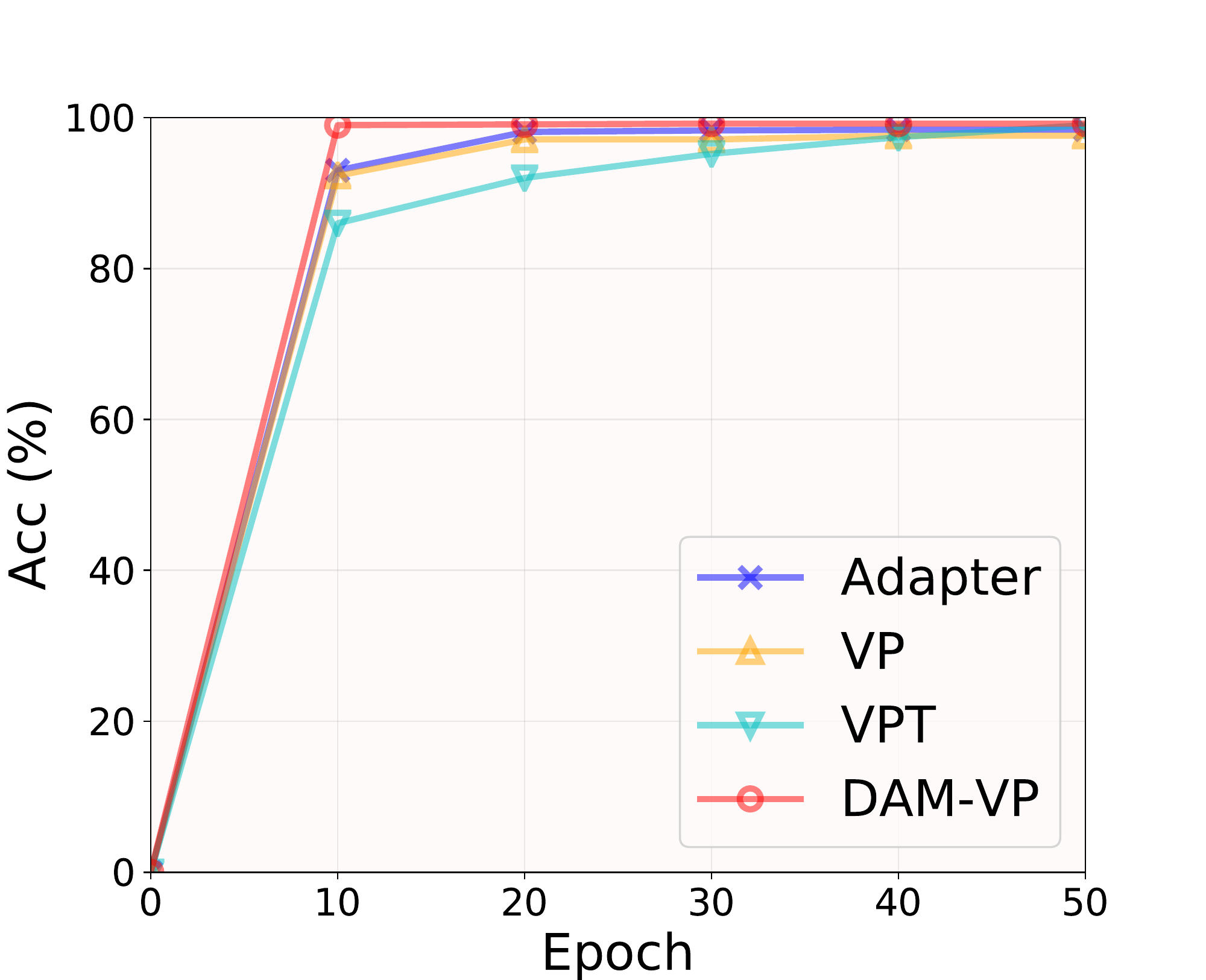}
    \\
    \footnotesize Oxford-Flowers
\end{minipage}
\hfill
\begin{minipage}{0.245\linewidth}
    \centering
    \includegraphics[width=1\linewidth]{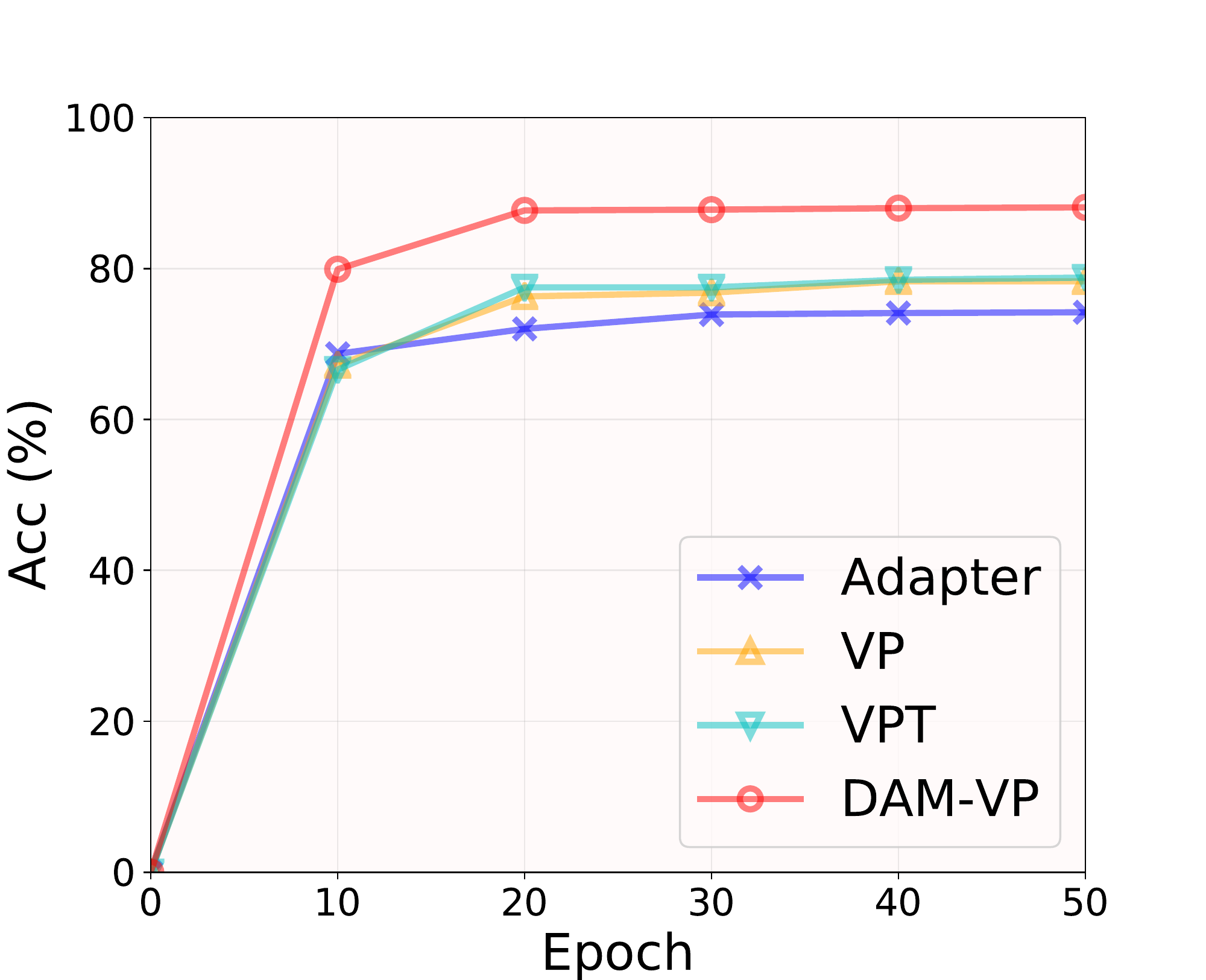}
    \\
    \footnotesize CIFAR100
\end{minipage}
\hfill
\begin{minipage}{0.245\linewidth}
    \centering
    \includegraphics[width=1\linewidth]{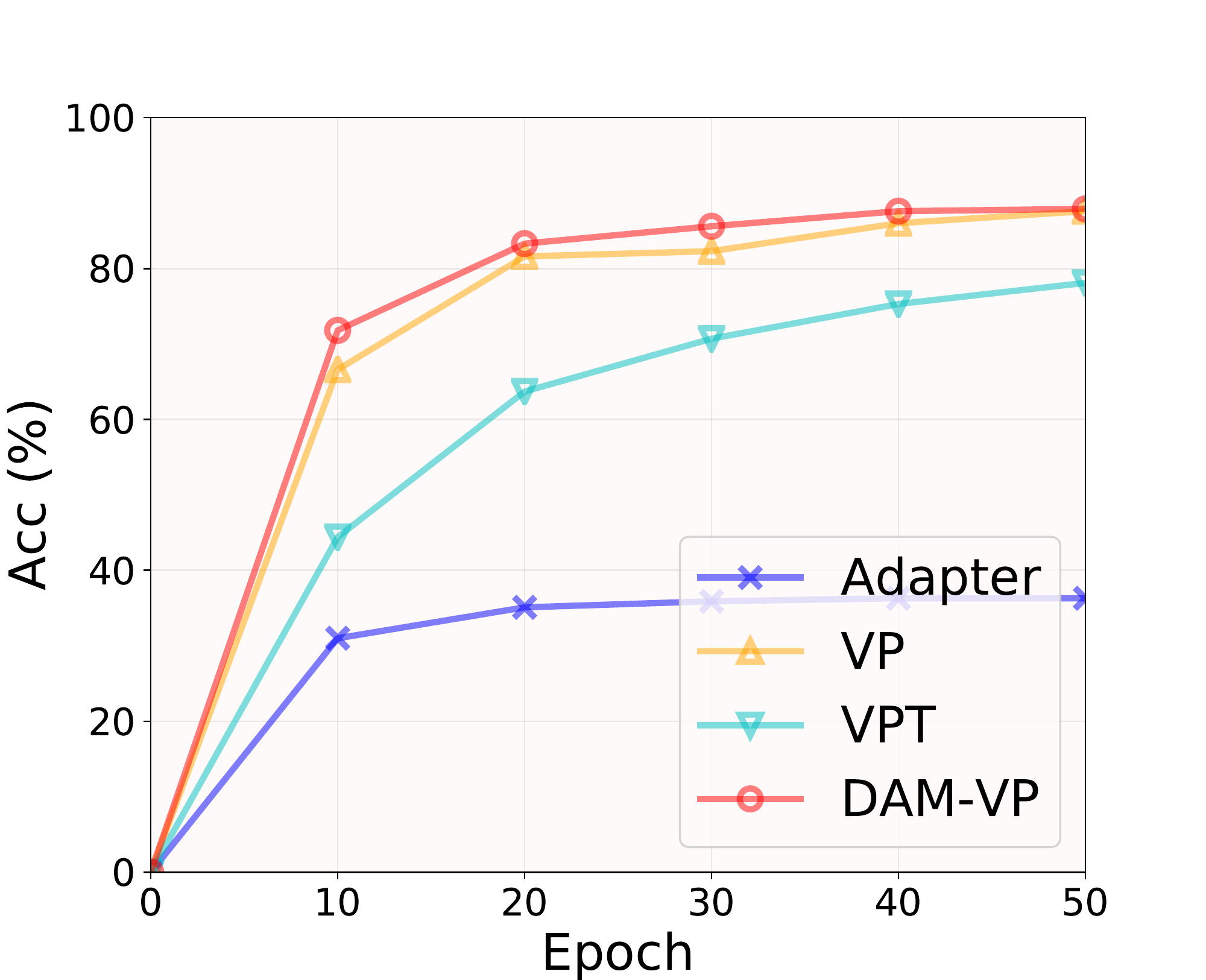}
    \\
    \footnotesize SVHN
\end{minipage}
\caption{Test accuracy curves of different adaption methods when adapting pre-trained ViT-B-1K to different datasets in head-freezing/missing scenario (\textbf{top row}) and adapting pre-trained ViT-B-22K to different datasets head-tuning scenario (\textbf{bottom row}).}
\label{fig:curve}
\end{figure*}

\noindent\textbf{Qualitative results.} 
To better observe the adaption performance of each method, we depict the Top-1 accuracy curve of the first 50 training epochs to investigate the differences. 
\Fref{fig:curve} shows the curve results of DAM-VP and the other three baselines in both head-freezing/missing and head-tuning scenarios, where four datasets with different diversities are selected. 
It is obvious that, the performance of DAM-VP is far ahead in the early stage, especially in the first 10 epochs. 
This phenomenon indicates our diversity-aware strategy can boost the efficiency of prompt optimization, since each prompt just need to learn from a group of images that already have considerable homogeneity. 
The design of meta-prompt initialization also benefits this quick converging of our method with a good start point.

\begin{table*}[t]
    \scriptsize
    \centering
    \setlength{\tabcolsep}{1.2mm}{
    \begin{tabular}{l|p{11mm}<{\centering}|p{9mm}<{\centering}p{9mm}<{\centering}p{9mm}<{\centering}p{9mm}<{\centering}p{9mm}<{\centering}p{9mm}<{\centering}p{11mm}<{\centering}p{11mm}<{\centering}p{9mm}<{\centering}p{9mm}<{\centering}|p{11mm}<{\centering}}
        \toprule
        & \multirow{1}{*}{Extra}
        & DTD
        & CUB200 
        & NABirds 
        & Dogs 
        & Flowers
        & Food101 
        & CIFAR100
        & CIFAR10 
        & GTSRB 
        & SVHN
        & Average
        \\
        & \multirow{1}{*}{Head}
        & \cite{dtd}
        & \cite{cub200}
        & \cite{nabirds}
        & \cite{stanford_dogs}
        & \cite{oxford_flowers}
        & \cite{food101}
        & \cite{cifar}
        & \cite{cifar}
        & \cite{gtsrb}
        & \cite{svhn}
        &
        \\
        \midrule
        Data diversity & - & 78.7 & 76.0 & 74.8 & 73.4 & 72.7 & 72.7 & 70.9 & 70.2 & 67.5 & 61.8 & -
        \\
        \midrule
        Fully-Tuning & \Checkmark & 64.3 & 87.3 & 82.7 & 89.4 & 98.8 & 84.9 & 68.9 & 97.4 & 97.1 & 87.4  & 85.8
        \\
        Linear & \Checkmark & 63.2 & 85.3 & 75.9 & 86.2 & 97.9 & 84.4 & 63.4 & 96.3 & 68.0 & 36.6 & 75.7
        \\
        Adapter \cite{adapter1,adapter2} & \Checkmark & 62.7 & 87.1 & 84.3 & 89.8 & 98.5 & 86.0 & 74.2 & 97.7 & 91.1 & 36.3 & 80.8
        \\
        VP \cite{bahng2022vp} & \Checkmark & 59.5 & 84.6 & 77.7 & 84.5 & 97.7 & 80.5 & 78.7 & 94.2 & 89.4 & 87.6 & 83.4
        \\
        VPT \cite{jia2022vpt} & \Checkmark & 65.8 & 88.5 & 84.2 & 90.2 & 99.0 & 83.3 & 78.8 & 96.8 & 90.7 & 78.1 & 85.5
        \\
        \cellcolor{gray!20}\textbf{DAM-VP (10 epochs)} & 
        \cellcolor{gray!20}\Checkmark &
        \cellcolor{gray!20}72.4 & \cellcolor{gray!20}86.3 & \cellcolor{gray!20}81.5 & \cellcolor{gray!20}92.2 & \cellcolor{gray!20}98.6 & \cellcolor{gray!20}86.2 & \cellcolor{gray!20}80.1 & \cellcolor{gray!20}90.5 & \cellcolor{gray!20}87.8 &  \cellcolor{gray!20}81.1 & \cellcolor{gray!20}85.7
        \\
        \cellcolor{gray!20}\textbf{DAM-VP (50 epochs)} & 
        \cellcolor{gray!20}\Checkmark & 
        \cellcolor{gray!20}73.1 &
        \cellcolor{gray!20}87.5 & \cellcolor{gray!20}82.1 & \cellcolor{gray!20}92.3 & \cellcolor{gray!20}99.2 & \cellcolor{gray!20}86.9 & \cellcolor{gray!20}88.1 & \cellcolor{gray!20}97.3 & \cellcolor{gray!20}90.6 &  \cellcolor{gray!20}87.9 & \cellcolor{gray!20}\textbf{88.5}
        \\
        \bottomrule
    \end{tabular}
    }
    \caption{Head-tuning adaption performance of different methods on ViT-B-22K, where we report image classification accuracy and all of baseline methods are trained for \textbf{100 epochs}.}
    \label{tab:table3}
\end{table*}

\begin{table*}[t]
    \scriptsize
    \centering
    \setlength{\tabcolsep}{1.2mm}{
    \begin{tabular}{l|p{11mm}<{\centering}|p{9mm}<{\centering}p{9mm}<{\centering}p{9mm}<{\centering}p{9mm}<{\centering}p{9mm}<{\centering}p{9mm}<{\centering}p{11mm}<{\centering}p{11mm}<{\centering}p{9mm}<{\centering}p{9mm}<{\centering}|p{11mm}<{\centering}}
        \toprule
        & \multirow{1}{*}{Extra}
        & DTD
        & CUB200 
        & NABirds 
        & Dogs 
        & Flowers
        & Food101 
        & CIFAR100
        & CIFAR10 
        & GTSRB 
        & SVHN
        & Average
        \\
        & \multirow{1}{*}{Head}
        & \cite{dtd}
        & \cite{cub200}
        & \cite{nabirds}
        & \cite{stanford_dogs}
        & \cite{oxford_flowers}
        & \cite{food101}
        & \cite{cifar}
        & \cite{cifar}
        & \cite{gtsrb}
        & \cite{svhn}
        &
        \\
        \midrule
        Data diversity & - & 78.7 & 76.0 & 74.8 & 73.4 & 72.7 & 72.7 & 70.9 & 70.2 & 67.5 & 61.8 & -
        \\
        \midrule
        Fully-Tuning & \Checkmark & 72.4 & 89.7 & 86.8 & 86.2 & 98.3 & 91.7 & 73.3 & 98.3 & 97.1 & 91.2 & 88.5
        \\
        Linear & \Checkmark & 73.6 & 88.6 & 85.2 & 85.9 & 99.4 & 88.2 & 61.6 & 96.3 & 83.8 & 43.5 & 80.6
        \\
        Adapter \cite{adapter1,adapter2} & \Checkmark & 73.9 & 88.5 & 84.6 & 86.8 & 98.9 & 88.7 & 85.7 & 96.5 & 83.6 & 71.3 & 85.9
        \\
        VP \cite{bahng2022vp} & \Checkmark & 75.1 & 86.5 & 82.9 & 81.3 & 98.6 & 83.4 & 80.6 & 94.8 & 82.4 & 80.3 & 84.6
        \\
        VPT \cite{jia2022vpt} & \Checkmark & 78.5 & 90.0 & 85.4 & 84.8 & 99.3 & 90.1 & 80.5 & 96.9 & 86.2 & 87.8 & 87.9
        \\
        \cellcolor{gray!20}\textbf{DAM-VP (10 epochs)} & 
        \cellcolor{gray!20}\Checkmark &
        \cellcolor{gray!20}77.0 & \cellcolor{gray!20}89.4 & \cellcolor{gray!20}86.8 & \cellcolor{gray!20}88.3 & \cellcolor{gray!20}99.6 & \cellcolor{gray!20}90.2 & \cellcolor{gray!20}85.5 & \cellcolor{gray!20}96.4 & \cellcolor{gray!20}84.7 &  \cellcolor{gray!20}79.0 &  \cellcolor{gray!20}87.7
        \\
        \cellcolor{gray!20}\textbf{DAM-VP (50 epochs)} & 
        \cellcolor{gray!20}\Checkmark & 
        \cellcolor{gray!20}80.0 &
        \cellcolor{gray!20}90.4 & \cellcolor{gray!20}86.9 & \cellcolor{gray!20}88.5 & \cellcolor{gray!20}99.6 & \cellcolor{gray!20}90.5 & \cellcolor{gray!20}88.1 & \cellcolor{gray!20}97.3 & \cellcolor{gray!20}86.8 &  \cellcolor{gray!20}81.7 &  \cellcolor{gray!20}\textbf{89.0}
        \\
        \bottomrule
    \end{tabular}
    }
    \caption{Head-tuning adaption performance of different methods on Swin-B-22K, where we report image classification accuracy and all of baseline methods are trained for \textbf{100 epochs}.}
    \label{tab:table4}
\end{table*}

\subsection{Ablation Study} 

\noindent\textbf{Component ablation.} 
We first verify the significance of the proposed diversity-aware strategy and the meta-prompt initialization on four aforementioned datasets that have different diversities. 
As listed in \Tref{tab:abl_1}, both two components contribute a lot to boosting prompting performance, especially when dealing with task data that has high diversity.

\begin{figure}[t]
\centering
\vspace{-2em}
\begin{minipage}{0.495\linewidth}
    \centering
    \includegraphics[width=1\linewidth]{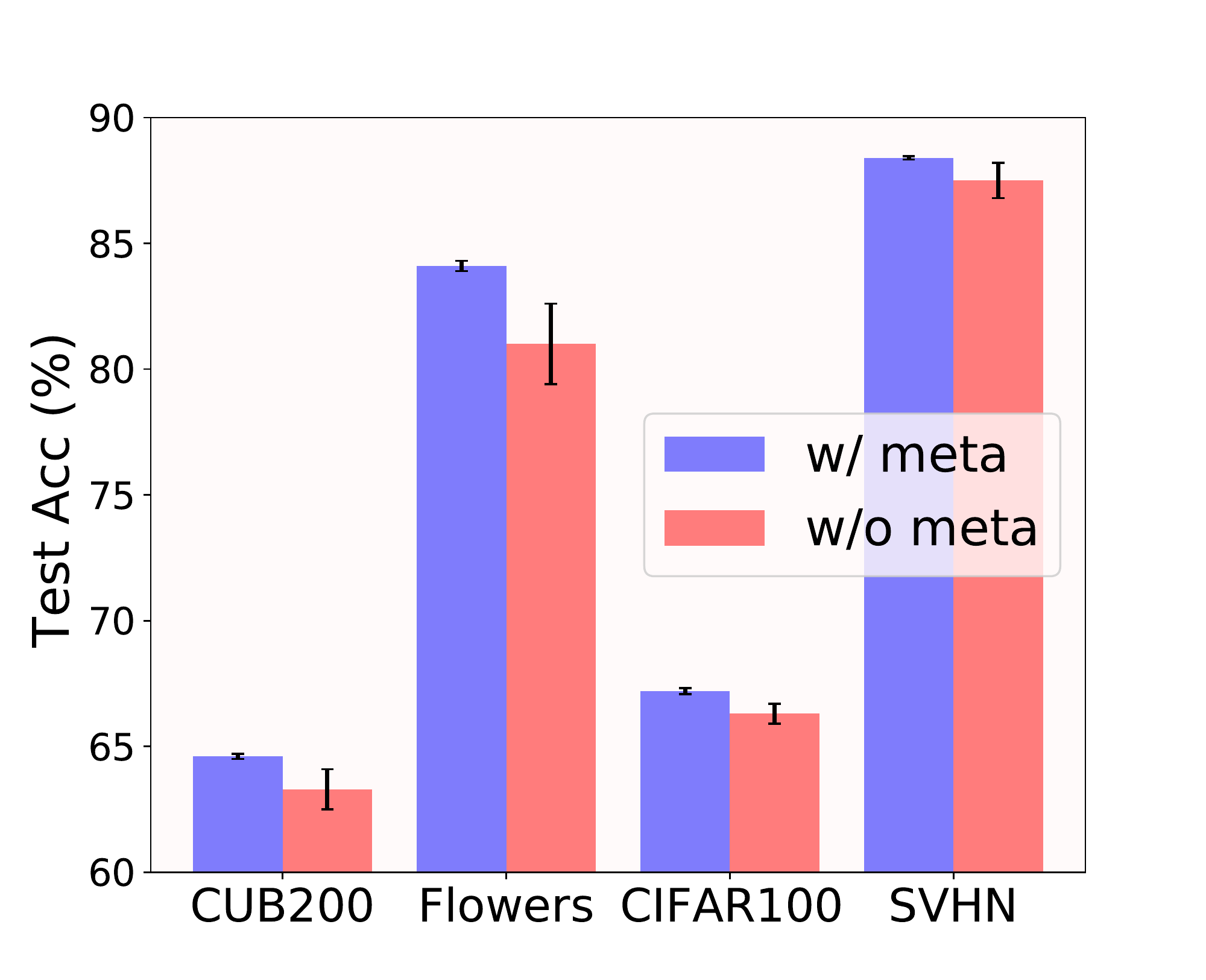}
\end{minipage}
\hfill
\begin{minipage}{0.495\linewidth}
    \centering
    \includegraphics[width=1\linewidth]{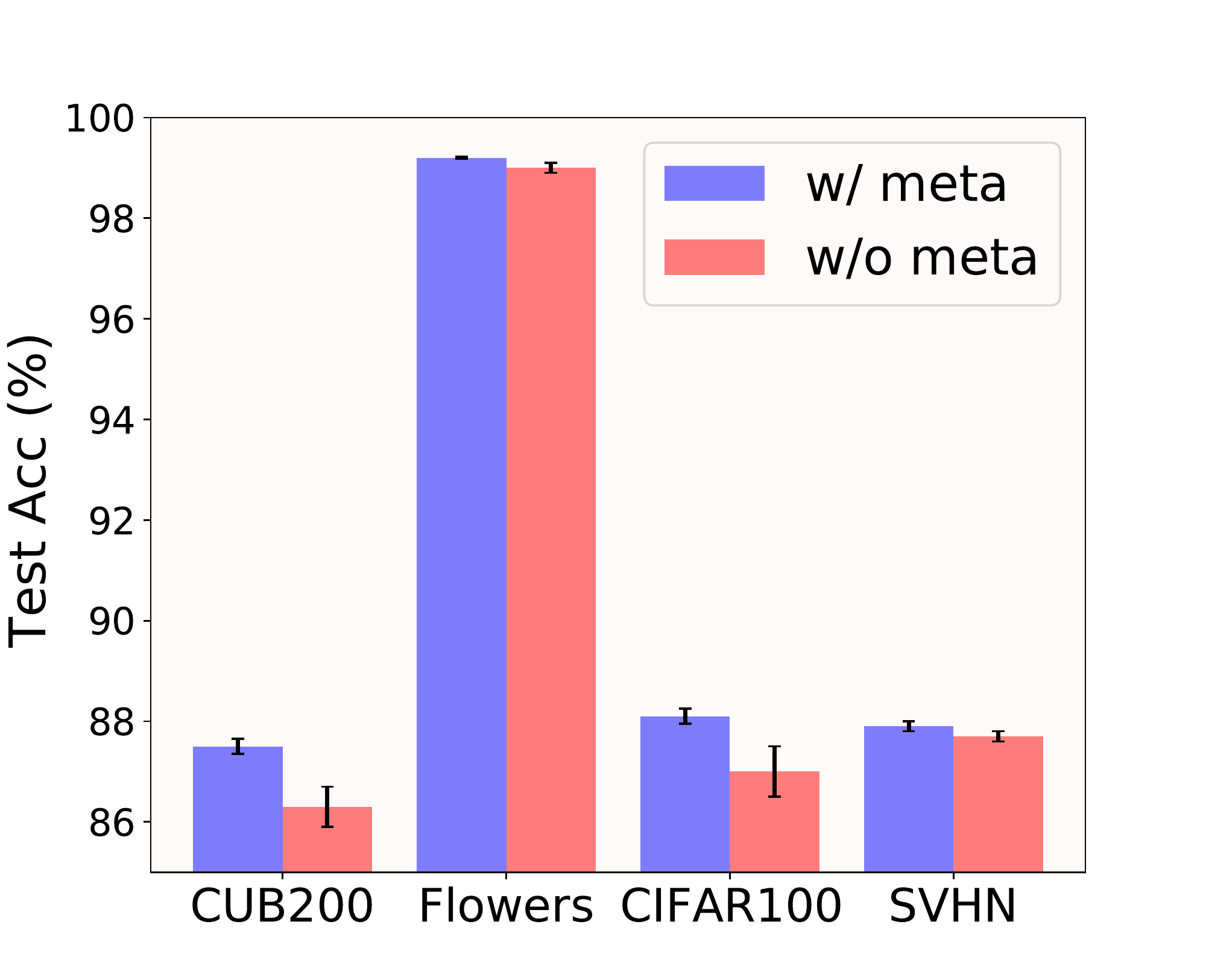}
\end{minipage}
\caption{Meta-prompt initialization makes prompt tuning more robust to the random factor. We test in both head-freezing/missing (ViT-B-1K, \textbf{left}) and head-tuning (ViT-B-22K, \textbf{right}) cases.}
\label{fig:random_vs_meta}
\end{figure}

\noindent\textbf{Prompt learning stability.} 
Previous methods mainly initialize the prompt randomly, this leads to the instability of the performance caused by different random seeds.
On the contrary, our meta-prompt design provides a good initialization point for optimization, which not only boosts the prompting performance but also improves the training stability.
To eliminate the impact of random seeds, we test on 5 random seeds in experiments and report the results in \Fref{fig:random_vs_meta}. 
It can be found that our meta-prompt initialization provides superb robustness across different training random seeds, while the randomly initialized prompt is unstable.

\begin{figure}[t]
\centering
\vspace{-2em}
\begin{minipage}{0.495\linewidth}
    \centering
    \includegraphics[width=1\linewidth]{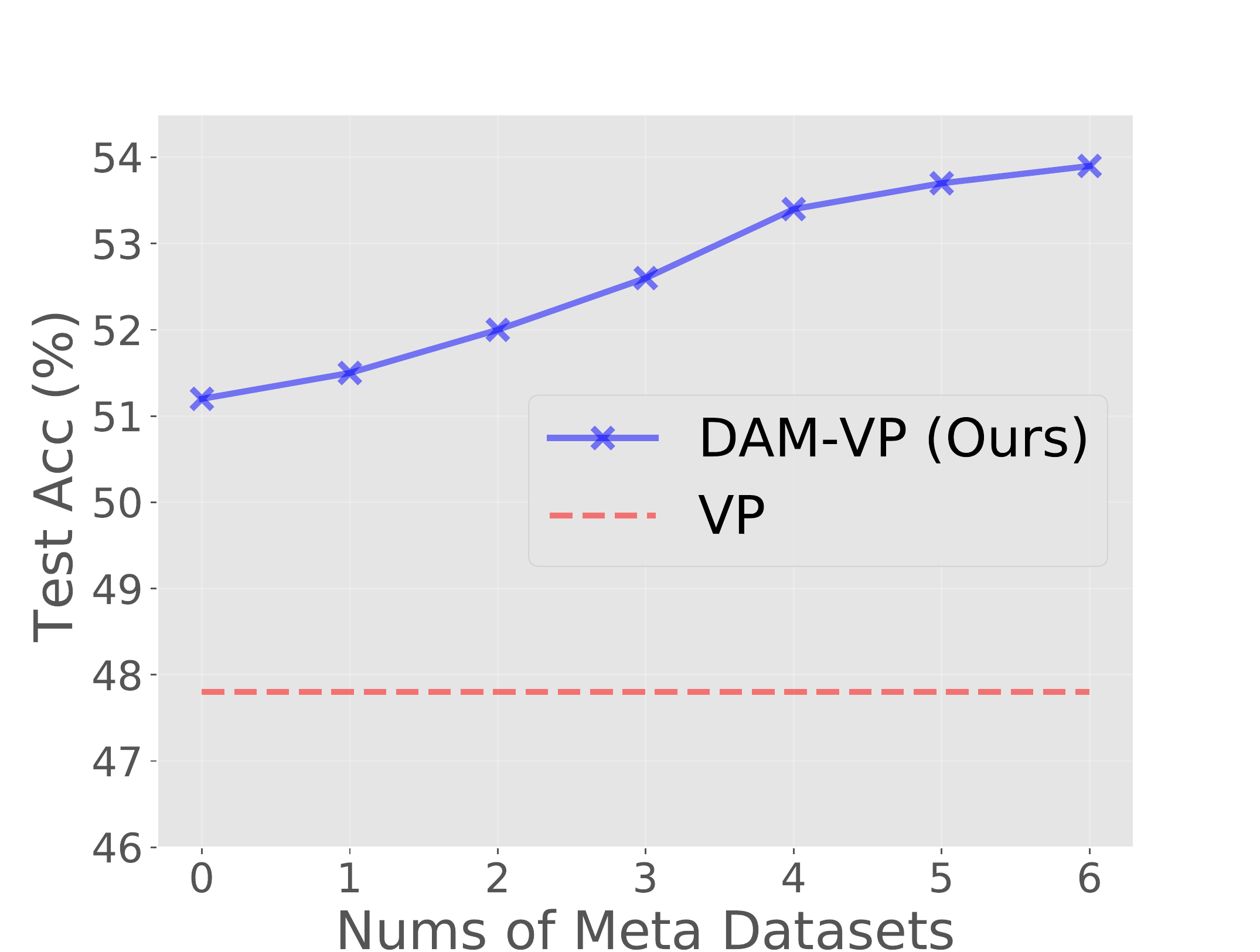}
\end{minipage}
\hfill
\begin{minipage}{0.495\linewidth}
    \centering
    \includegraphics[width=1\linewidth]{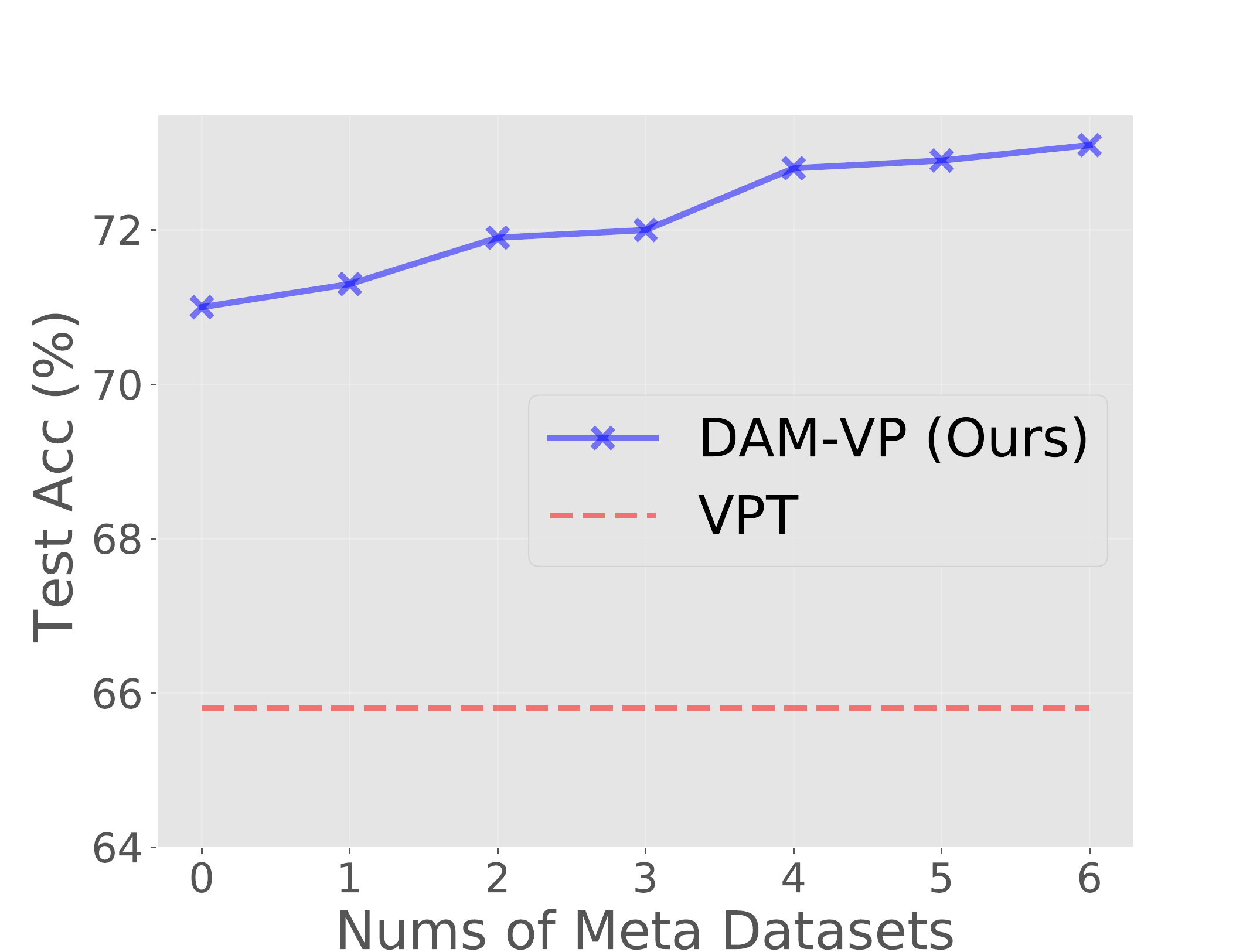}
\end{minipage}
\caption{Ablating meta-prompt dataset number on DTD in the head-freezing/missing setting (ViT-B-1K, \textbf{left}) and the head-tuning (ViT-B-22K, \textbf{right}) setting.}
\label{fig:abl_meta_dataset}
\end{figure}

\noindent\textbf{Meta-prompt dataset number.} 
As we elaborated in \sref{sec:setup}, there are 6 datasets prepared for our meta-prompt learning, which is also the default setting we used. 
Here we ablate this setting by reducing the number of meta datasets. 
As shown in \Fref{fig:abl_meta_dataset}, when the meta dataset number is 0, \ie, no meta-prompt is used, our DVM-VP gets a reasonable result that performs better than previous baseline methods.
With the number of meta datasets increases, the prompting performance in both head-freezing/missing and head-tuning scenarios generally gets boosted. 
It proves that more prompting knowledge obtained from previous data is quite helpful for visual prompts to reduce the data distribution gap between downstream tasks and pretraining tasks.

\begin{figure}[t]
\centering
\vspace{-2em}
\begin{minipage}{0.495\linewidth}
    \centering
    \includegraphics[width=1\linewidth]{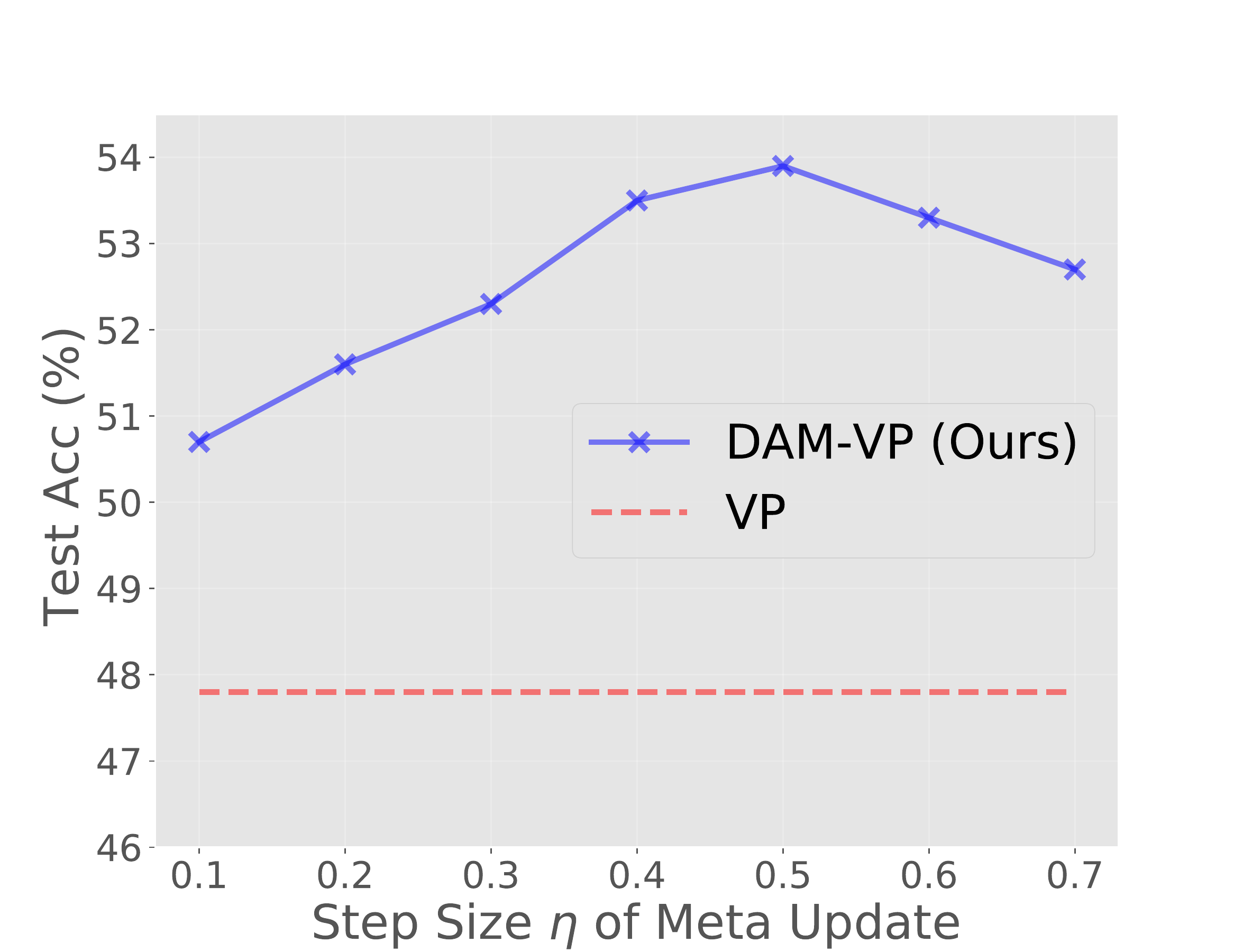}
\end{minipage}
\hfill
\begin{minipage}{0.495\linewidth}
    \centering
    \includegraphics[width=1\linewidth]{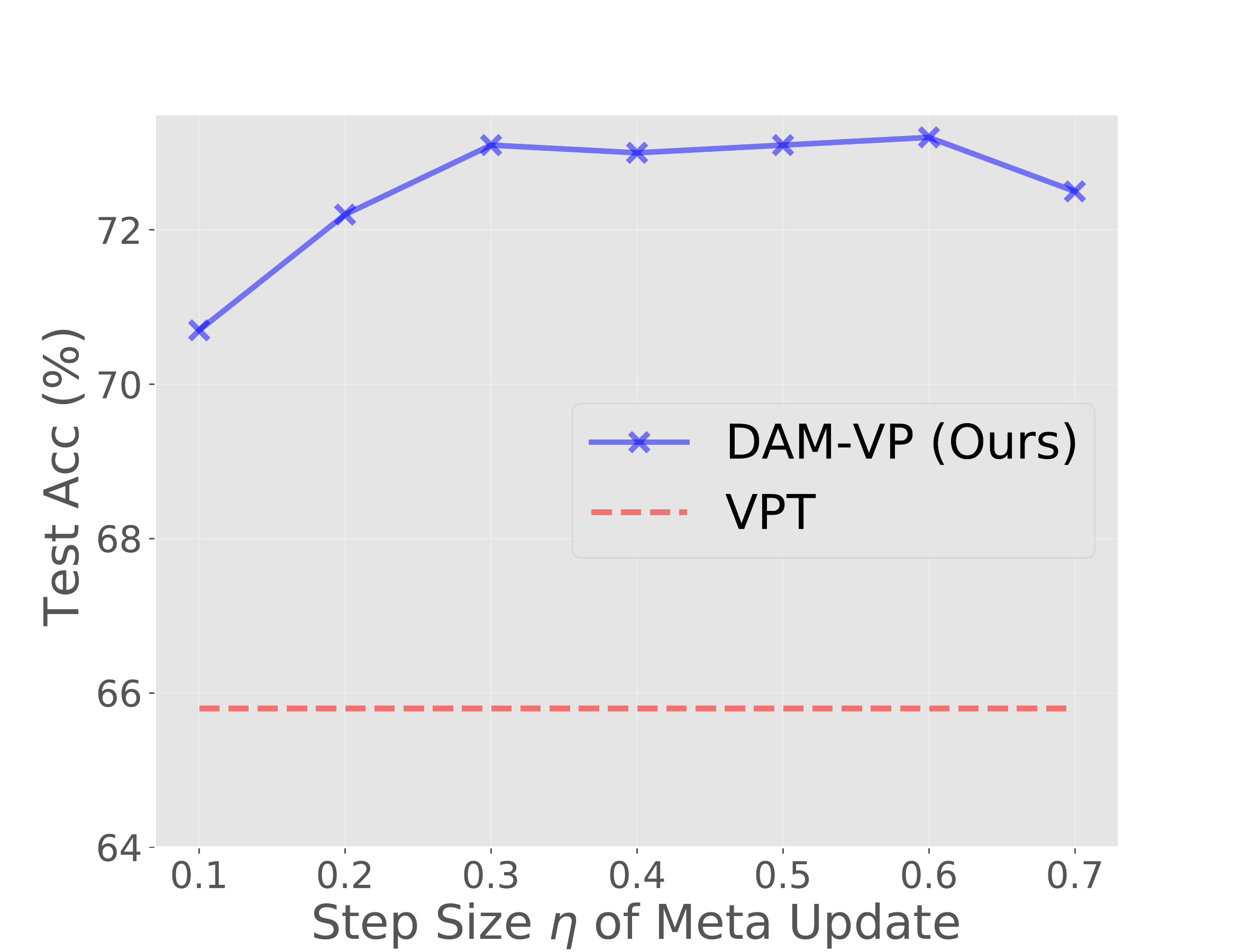}
\end{minipage}
\caption{Ablating step size $\eta$ on DTD in head-freezing/missing (ViT-B-1K, \textbf{left}) and head-tuning (ViT-B-22K, \textbf{right}) cases.}
\label{fig:abl_meta_update}
\end{figure}

\noindent\textbf{Meta-prompt update step size $\eta$.} 
The step size $\eta$ is a crucial hyper-parameter applied in the meta-prompt update. 
Here we ablate different step sizes that vary from 0.1 to 0.7 with the results given in \Fref{fig:abl_meta_update}. 
Compared with baselines, the performance is relatively robust to different step sizes. 
We choose $\eta=0.5$ as the default configuration.

\begin{figure}[t]
\vspace{-2em}
\centering
\begin{minipage}{0.495\linewidth}
    \centering
    \includegraphics[width=1\linewidth]{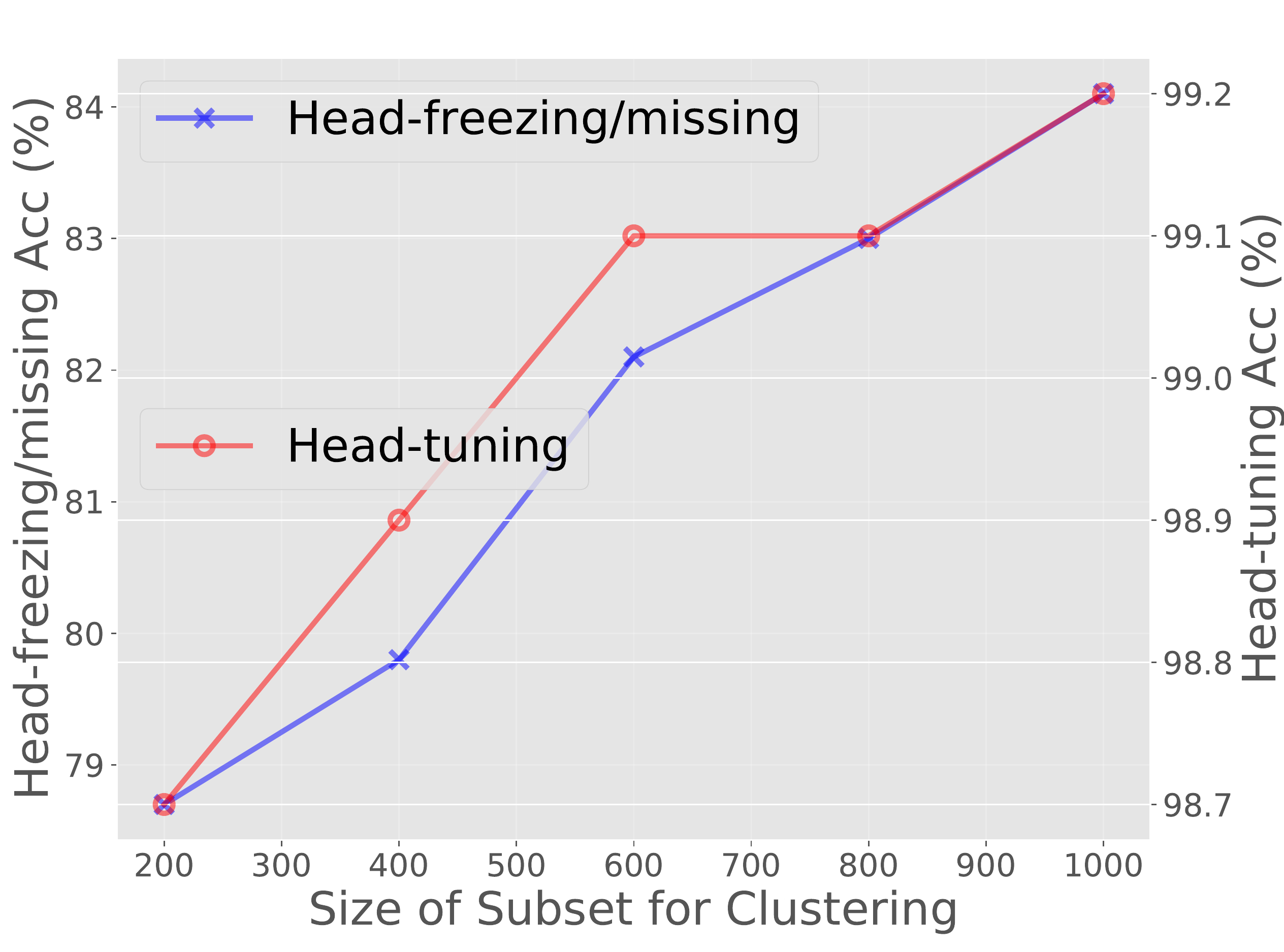}
\end{minipage}
\hfill
\begin{minipage}{0.495\linewidth}
    \centering
    \includegraphics[width=1\linewidth]{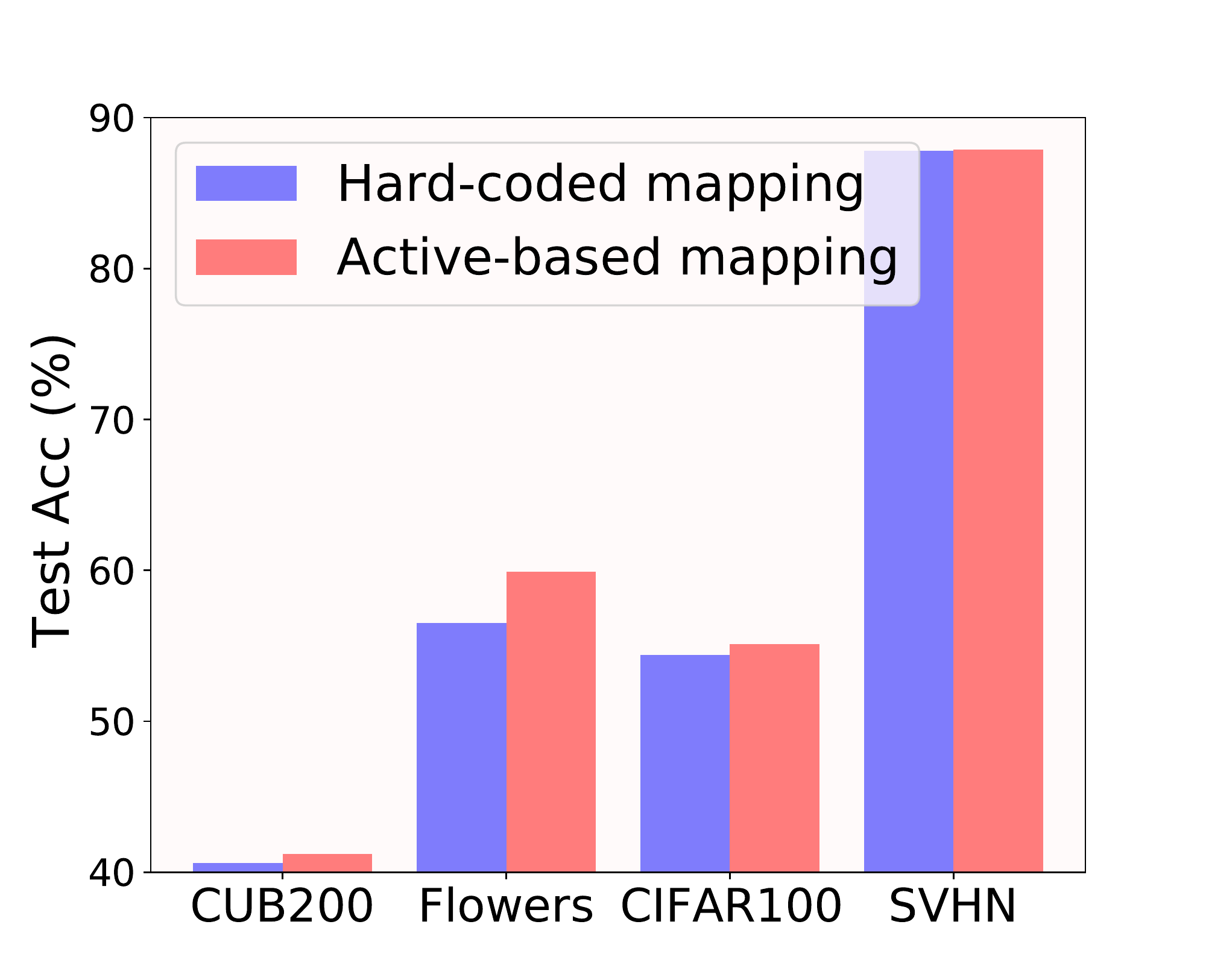}
\end{minipage}
\caption{(\textbf{Left}) Ablating the size of subset $\mathcal{S}_\mathcal{T}$ in both head-freezing/missing (ViT-B-1K) and head-tuning (ViT-B-22K) cases. (\textbf{Right}) Comparison between hard-coded mapping and our active-based mapping used for head-freezing/missing case.}
\label{fig:abl_meta_2}
\vspace{-0.5em}
\end{figure}

\noindent\textbf{Different size of subset $\mathcal{S}_\mathcal{T}$.}  
The size of subset $\mathcal{S}_\mathcal{T}$ can directly influence the clustering result and further affects the partition result during our diversity-adaptive dataset partition. 
Therefore, we apply different subset size that varies from 200 to 1000 to explore how it affects the prompting ability. 
From \Fref{fig:abl_meta_2} (left), we find that as the size of $\mathcal{S}_\mathcal{T}$ increases, we can get higher performance in both scenarios. Meanwhile, we find our divide-and-conquer design is quite efficient that even with only 200 images, our DAM-VP still outperforms previous methods by a large margin.
Considering the training images of some datasets are limited, we adopt 1,000 as the default size of $\mathcal{S}_\mathcal{T}$ in this work. 

\noindent\textbf{Hard-coded mapping \vs Active-based mapping}. 
As we discussed in \sref{sec:method}, we argue the hard-coded mapping used by \cite{bahng2022vp} for head-freezing/missing scenario is inefficient, since it might optimize some not active enough channels of the output feature~(\ie, relatively robust to diverse model input), thus we propose active-based mapping to alleviate this issue. 
\Fref{fig:abl_meta_2} (right) show the comparison between these two mapping methods on VP, tested on ViT-B-1K.

\section{Conclusion}

This paper considers the data diversity property in downstream task adaption for prompting pre-trained vision models. 
We argue that the per-dataset generic prompt adopt in previous methods can hardly handle the dataset of large data diversity. 
To address this, we propose DAM-VP based on diversity-adaptive dataset partition and prompt selection, where our prompts is initialized by a meta-prompt that learns through a quick meta learning algorithm.  
Extensive experiments prove the superior performance of DAM-VP in both head-freezing/missing and head-tuning cases.

\noindent\textbf{Acknowledgement.} This work was supported in part by the Natural Science Foundation of China under Grant U20B2047, 62072421, 62002334, 62102386 and 62121002, Key Research and Development program of Anhui Province under Grant 2022k07020008, the Fundamental Research Funds for the Central Universities under Grant WK5290000003, Ant Group through CCF-Ant Innovative Research Program CCF-AFSG RF20210025, Alibaba Group through Alibaba Innovative Research Program. This work was also partly supported by Shenzhen Key Laboratory of Media Security, Shenzhen University, Shenzhen 518060, China.

{\small
\bibliographystyle{ieee_fullname}
\bibliography{egbib}
}
\clearpage
\section{Supplementary Materials}
We provide supplementary materials for our paper ``Diversity-Aware Meta Visual Prompting'', including: 
\begin{itemize}
	\item Discussion on limitations and boarder impacts. 
	
	\item Discussion on tunable parameters. 
	
	\item Setting for models w/o task-specific heads.
	
	\item Setting for convolution networks.
	
	\item More details regarding datasets. 
	
	\item More details regarding model backbones. 
	
	\item More results on MoCo-v3 and ResNet-50. 
	
	\item More results on VTAB-1k benchmark.
	
	\item More details regarding hyper-parameters. 
	
	\item Ablation study on clustering threshold.
	
	
\end{itemize}

Each part is specified as follows, respectively.

\vspace{1em}
\large\noindent\textbf{A. Limitations and Social Impacts}
\vspace{0.5em}
\normalsize

Here we discuss the shortcomings and the potential social influences of the proposed diversity-aware meta visual prompting (DAM-VP), respectively. 

\textbf{For limitations}, two aspects of concerns might be raised. 
First, it is obvious that DAM-VP introduces more visual prompts than VP \cite{bahng2022vp} which trains the universal task-specific prompt. 
At the first glance, learning multiple visual prompts on a particular downstream task seems less parameter-efficient during adaption. 
However, we should argue that the amount of prompts introduced by our method is quite reasonable, \eg, $\sim$25 for ViT-B-22K averaged on 10 datasets. 
This amount of extra tunable parameters brought by DAM-VP is less than that is brought by an additional linear head. 
The tunable parameters brought by DAM-VP is comparable with baselines methods, which is detailed in Sec. B and showcased in \Tref{tab:params}. 
Relative to tuning all of pre-trained model parameters, the amount of extra tunable parameters brought by our method is really insignificant, which has very limited threat to the storage. 
On the other hand, when the number of tunable parameters introduced is small enough, it makes no sense to compare the efficiency of different methods only by comparing the number of tunable parameters. 
\textbf{We should claim that the efficiency of our method is mainly reflected in our ability to converge faster, \eg, using 10 epochs to be comparable with (or even surpass) the performances of baselines that trains for 100 epochs.} 

\textbf{For social impacts}, it is clear that exploring more effective and efficient visual prompting methods can greatly benefit the adaption of nowadays huge pre-trained models on downstream tasks. Visual prompting provides a novel perspective for boosting transfer learning performance of pre-trained vision models. 
It is crucial, at least on the aspect of application, for pre-trained models that has large capacity and capability to be easily re-programmed in both industry and academia.

\begin{table}[t]
	\footnotesize
	\centering
	\setlength{\tabcolsep}{1mm}{
		\begin{tabular}{lp{9mm}<{\centering}p{9mm}<{\centering}p{9mm}<{\centering}p{9mm}<{\centering}p{9mm}<{\centering}p{9mm}<{\centering}}
			\toprule
			& FT 
			& LP
			& Adapter
			& VP
			& VPT
			& Ours
			\\
			\midrule
			Total params & $10.01\times$ & $0.43\times$ & $0.51\times$ & $0.44\times$ & $0.49\times$ & $0.63\times$
			\\
			\bottomrule
		\end{tabular}
	}
	\caption{Total tunable parameters needed for 10 datasets when adapting ViT-B-22K in the head-tuning scenario, where ``$\times$'' the multiple of the amount of tunable parameters relative to the total amount of pre-trained ViT-B-22K encoder parameters ($\sim$85.8M). Here ``FT'' means fully-tuning and ``LP'' means linear probing.}
	\label{tab:params}
\end{table}

\begin{table*}[t]
	\footnotesize
	\centering
	\setlength{\tabcolsep}{1.2mm}{
		\begin{tabular}{lp{16mm}<{\centering}p{14mm}<{\centering}p{15mm}<{\centering}p{11mm}<{\centering}p{11mm}<{\centering}p{11mm}<{\centering}p{11mm}<{\centering}p{11mm}<{\centering}}
			\toprule
			Dataset
			& Usage
			& Meta Class
			& \# Categories
			& Train 
			& Val 
			& Test 
			& Diversities
			& Prompts 
			\\
			\midrule
			DTD \cite{dtd}
			& \multirow{10}{*}{Evaluation} 
			& textures
			& 47
			& 1,880 
			& 1,880 
			& 1,880 
			& 78.7 
			& 154 
			\\
			CUB200 \cite{cub200}
			& 
			& birds 
			& 200 
			& 5,394 
			& 600 
			& 5,794 
			& 76.0
			& 18
			\\
			NABirds \cite{nabirds}
			& 
			& birds
			& 555 
			& 21,536	
			& 2,393 
			& 24,633 
			& 74.8 
			& 22 
			\\
			Stanford-Dogs \cite{stanford_dogs}
			& 
			& dogs 
			& 120 
			& 10,800	
			& 1,200 
			& 8,580 
			& 73.4 
			& 33 
			\\
			Oxford-Flowers \cite{oxford_flowers}
			& 
			& flowers 
			& 102 
			& 1,020 
			& 1,020 
			& 6,149 
			& 72.7 
			& 26 
			\\
			Food101 \cite{food101}
			& 
			& food dishes 
			& 101 
			& 60,600 
			& 15,150 
			& 25,250 
			& 72.7 
			& 51
			\\
			CIFAR100 \cite{cifar}
			& 
			& all 
			& 100 
			& 40,000 
			& 10,000
			& 10,000
			& 70.9
			& 79 
			\\
			CIFAR10 \cite{cifar}
			& 
			& all 
			& 10 
			& 40,000 
			& 10,000
			& 10,000
			& 70.2
			& 42
			\\
			GTSRB \cite{gtsrb}
			& 
			& traffic signs
			& 43 
			& 21,312 
			& 2,526 
			& 12,630 
			& 67.5
			& 6
			\\
			SVHN \cite{svhn}
			& 
			& numbers
			& 10
			& 58,605 
			& 14,652 
			& 26,032 
			& 61.8 
			& 3
			\\
			\midrule
			SUN397 \cite{sun397}
			& \multirow{6}{*}{Meta Training} 
			& scenes
			& 397
			& 108,754
			& -
			& -
			& 76.9
			& 128
			\\
			STL10 \cite{stl10}
			& 
			& all
			& 10 
			& 5,000
			& -
			& 8,000
			& 74.1
			& 43
			\\
			Fru92 \cite{hou2017vegfru}
			& 
			& fruits
			& 92
			& 9,200
			& 4,600
			& 55,814
			& 74.1
			& 42
			\\
			Oxford-IIIT Pet \cite{parkhi2012cats}
			& 
			& cats,dogs
			& 37
			& 3,680
			& -
			& 3,669
			& 72.4
			& 18
			\\
			Veg200 \cite{hou2017vegfru}
			& 
			& vegetables
			& 200
			& 20,000
			& 10,000
			& 61,117
			& 71.5
			& 95
			\\
			EuroSAT \cite{helber2019eurosat}
			& 
			& remote
			& 10
			& 27,000
			& -
			& -
			& 64.6
			& 12
			\\
			\bottomrule
		\end{tabular}
	}
	\caption{Basic information of the datasets used in our work. ``Prompts'' shows the prompt numbers used on ViT-B-1K in the head-freezing/missing scenario.}
	\label{tab:dataset}
\end{table*}

\vspace{1em}
\large\noindent \textbf{B. Discussion on Tunable parameters}
\vspace{0.5em}
\normalsize

Although keeping the pre-trained models untouched, our visual prompts are also the extra introduced parameters for transfer learning. 
We compare the amount of tunable parameters of different methods on ViT-B-22K in the head-tuning scenario, showcased in \Tref{tab:params}. 
Apparently, our method DAM-VP uses the similar amount of tunable parameters with previous visual prompting methods, indicating the comparable parameter efficiency. 
Compared with VPT \cite{jia2022vpt}, the slightly more tunable parameters introduced by DAM-VP is relatively tolerable and acceptable since they are both far away less than FT. 
However, it can not reflect the efficiency during adaption. 
As we stated in limitations, our method is more efficient than other methods thanks to its faster converging, using 10 epochs to be comparable with (or even surpass) the previous methods that use 100 epochs.

\vspace{1em}
\large\noindent\textbf{C. Setting for Models w/o Task-Specific Heads}
\vspace{0.5em}
\normalsize

In the head-freezing/missing scenario, the task-specific is discarded so that it is necessary to design an approach to map the output feature to our desired classification logits. 
Previous VP \cite{bahng2022vp} applies a hard-coded mapping method to tackle with this, \ie, directly using the first $N_c$ channels of feature output as the classification probability output of $N_c$ categories. 
However, we argue that this method is too straightforward that it ignores the important property of neural networks, \ie, usually, some neurons in the intermediate layer might be not sufficiently active and relatively robust to the different inputs. 
This denotes that some of the selected feature channels selected by hard-coded mapping probably have very limited space for their variation, since their corresponding neurons are more ``robust''. 
In other words, the optimization of visual prompts might be seriously hindered by these less active channels.

To alleviate this issue, we propose active-based mapping, a simple but effective method for converting features to logits. 
Specifically, given a pre-trained vision encoder $\mathcal{M}$, we input it with a batch of randomly generated Gaussian noises to observe each channel's variance of the output visual feature. 
By sorting these variances, we can obtain the ranking of the sensitivities of output feature channels and select the largest $N_c$ channels as our desired active channels. 
After normalized, these $N_c$ channels can construct the output probabilities of any input image.

\vspace{1em}
\large\noindent\textbf{D. Setting for Convolution Networks}
\vspace{0.5em}
\normalsize

Different from previous methods such as VPT \cite{jia2022vpt} and Adapter \cite{adapter1,adapter2}, our method is universal for both Vision Transformer and convolution networks since our prompt design is consistent with VP \cite{bahng2022vp} that applies pixel-level visual prompts. 
The prompt is actually the learnable pixel patches, which looks like a photo frame with the width of 30 and can be added on the original image as input. 
We choose this design mainly because: 
1) it naturally suits all kinds of vision models since directly crafting pixels guarantees that only the input space is considered to be modified. 
2) The photo-frame-like structure can greatly inherent the main content of the input image, which usually allocates at the center of the image. 
In this supplementary, we also provide the prompting results on ResNet-50 \cite{he2016deep} that is pre-trained on ImageNet-1k in \Tref{tab:resnet}.

\begin{table}[t]
	\footnotesize
	\centering
	\setlength{\tabcolsep}{1mm}{
		\begin{tabular}{lccccc}
			\toprule
			\multirow{2}{*}{Name}
			& \multirow{2}{*}{Backbone} 
			& Pre-trained
			& Pre-trained
			& Params
			& Feature
			\\
			& 
			& Paradigm
			& Dataset
			& (M)
			& Dim
			\\
			\midrule
			ViT-B-1K
			& ViT-B/16
			& Supervised
			& ImageNet-1k
			& 85
			& 768
			\\
			ViT-B-22K
			& ViT-B/16
			& Supervised
			& ImageNet-22k
			& 85
			& 768
			\\
			CLIP-ViT-B
			& ViT-B/16
			& CLIP
			& 400M web data
			& 85
			& 768
			\\
			Swin-B-22K
			& Swin-B
			& Supervised
			& ImageNet-22k
			& 88
			& 1024
			\\
			MoCo-B-1K
			& ViT-B/16
			& Contrastive
			& ImageNet-1k
			& 85
			& 768
			\\
			ResNet50-1K
			& ResNet-50
			& Supervised
			& ImageNet-1k
			& 23
			& 2048
			\\
			\bottomrule
		\end{tabular}
	}
	\caption{Basic information of the pre-trained vision backbones used in our experiment.}
	\label{tab:backbones}
\end{table}

\begin{figure*}[t]
	\centering
	\begin{minipage}{0.01\linewidth}
		\centering
		\footnotesize \rotatebox{90}{DTD}
	\end{minipage}
	\hfill
	\begin{minipage}{0.11\linewidth}
		\centering
		\includegraphics[width=1\linewidth]{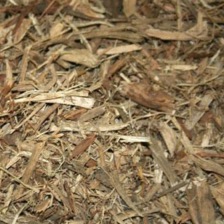}
	\end{minipage}
	\hfill
	\begin{minipage}{0.11\linewidth}
		\centering
		\includegraphics[width=1\linewidth]{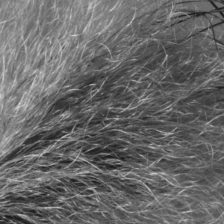}
	\end{minipage}
	\hfill
	\begin{minipage}{0.11\linewidth}
		\centering
		\includegraphics[width=1\linewidth]{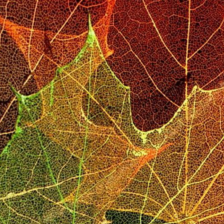}
	\end{minipage}
	\hfill
	\begin{minipage}{0.11\linewidth}
		\centering
		\includegraphics[width=1\linewidth]{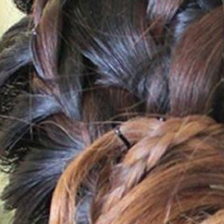}
	\end{minipage}
	\hfill
	\begin{minipage}{0.11\linewidth}
		\centering
		\includegraphics[width=1\linewidth]{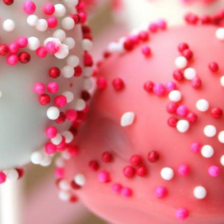}
	\end{minipage}
	\hfill
	\begin{minipage}{0.11\linewidth}
		\centering
		\includegraphics[width=1\linewidth]{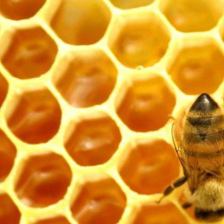}
	\end{minipage}
	\hfill
	\begin{minipage}{0.11\linewidth}
		\centering
		\includegraphics[width=1\linewidth]{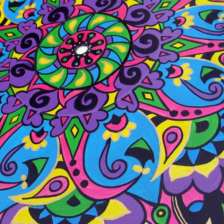}
	\end{minipage}
	\hfill
	\begin{minipage}{0.11\linewidth}
		\centering
		\includegraphics[width=1\linewidth]{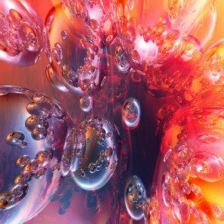}
	\end{minipage}
	\vfill
	\vspace{0.5em}
	\begin{minipage}{0.01\linewidth}
		\centering
		\footnotesize \rotatebox{90}{CUB200}
	\end{minipage}
	\hfill
	\begin{minipage}{0.11\linewidth}
		\centering
		\includegraphics[width=1\linewidth]{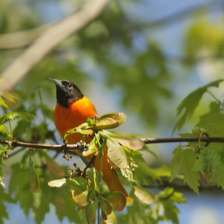}
	\end{minipage}
	\hfill
	\begin{minipage}{0.11\linewidth}
		\centering
		\includegraphics[width=1\linewidth]{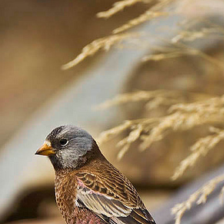}
	\end{minipage}
	\hfill
	\begin{minipage}{0.11\linewidth}
		\centering
		\includegraphics[width=1\linewidth]{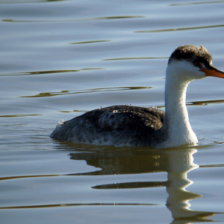}
	\end{minipage}
	\hfill
	\begin{minipage}{0.11\linewidth}
		\centering
		\includegraphics[width=1\linewidth]{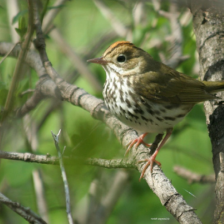}
	\end{minipage}
	\hfill
	\begin{minipage}{0.11\linewidth}
		\centering
		\includegraphics[width=1\linewidth]{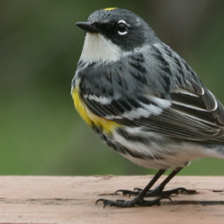}
	\end{minipage}
	\hfill
	\begin{minipage}{0.11\linewidth}
		\centering
		\includegraphics[width=1\linewidth]{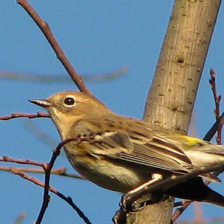}
	\end{minipage}
	\hfill
	\begin{minipage}{0.11\linewidth}
		\centering
		\includegraphics[width=1\linewidth]{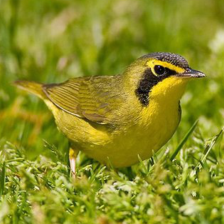}
	\end{minipage}
	\hfill
	\begin{minipage}{0.11\linewidth}
		\centering
		\includegraphics[width=1\linewidth]{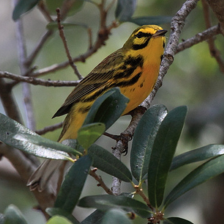}
	\end{minipage}
	\vfill
	\vspace{0.5em}
	\begin{minipage}{0.01\linewidth}
		\centering
		\footnotesize \rotatebox{90}{NABirds}
	\end{minipage}
	\hfill
	\begin{minipage}{0.11\linewidth}
		\centering
		\includegraphics[width=1\linewidth]{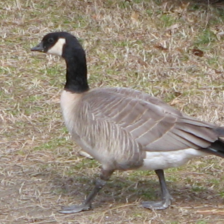}
	\end{minipage}
	\hfill
	\begin{minipage}{0.11\linewidth}
		\centering
		\includegraphics[width=1\linewidth]{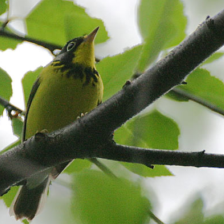}
	\end{minipage}
	\hfill
	\begin{minipage}{0.11\linewidth}
		\centering
		\includegraphics[width=1\linewidth]{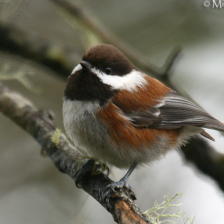}
	\end{minipage}
	\hfill
	\begin{minipage}{0.11\linewidth}
		\centering
		\includegraphics[width=1\linewidth]{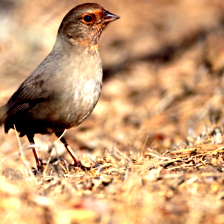}
	\end{minipage}
	\hfill
	\begin{minipage}{0.11\linewidth}
		\centering
		\includegraphics[width=1\linewidth]{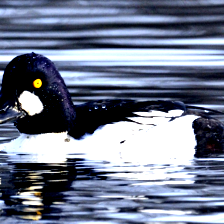}
	\end{minipage}
	\hfill
	\begin{minipage}{0.11\linewidth}
		\centering
		\includegraphics[width=1\linewidth]{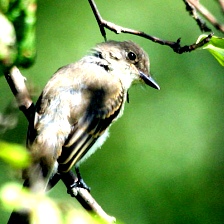}
	\end{minipage}
	\hfill
	\begin{minipage}{0.11\linewidth}
		\centering
		\includegraphics[width=1\linewidth]{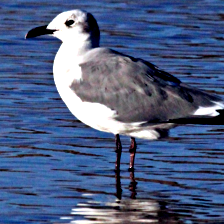}
	\end{minipage}
	\hfill
	\begin{minipage}{0.11\linewidth}
		\centering
		\includegraphics[width=1\linewidth]{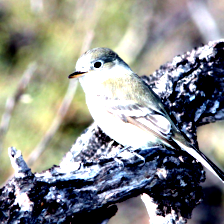}
	\end{minipage}
	\vfill
	\vspace{0.5em}
	\begin{minipage}{0.01\linewidth}
		\centering
		\footnotesize \rotatebox{90}{Stanford-Dogs}
	\end{minipage}
	\hfill
	\begin{minipage}{0.11\linewidth}
		\centering
		\includegraphics[width=1\linewidth]{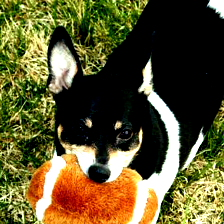}
	\end{minipage}
	\hfill
	\begin{minipage}{0.11\linewidth}
		\centering
		\includegraphics[width=1\linewidth]{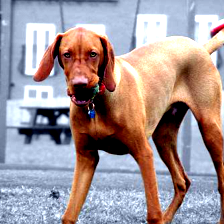}
	\end{minipage}
	\hfill
	\begin{minipage}{0.11\linewidth}
		\centering
		\includegraphics[width=1\linewidth]{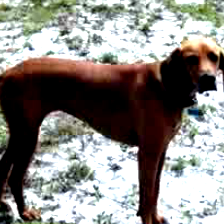}
	\end{minipage}
	\hfill
	\begin{minipage}{0.11\linewidth}
		\centering
		\includegraphics[width=1\linewidth]{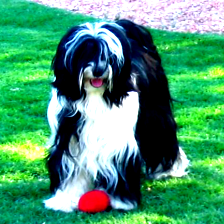}
	\end{minipage}
	\hfill
	\begin{minipage}{0.11\linewidth}
		\centering
		\includegraphics[width=1\linewidth]{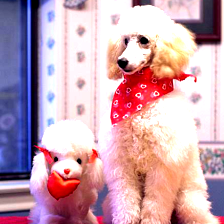}
	\end{minipage}
	\hfill
	\begin{minipage}{0.11\linewidth}
		\centering
		\includegraphics[width=1\linewidth]{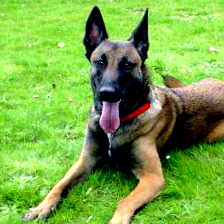}
	\end{minipage}
	\hfill
	\begin{minipage}{0.11\linewidth}
		\centering
		\includegraphics[width=1\linewidth]{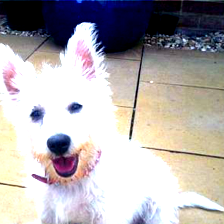}
	\end{minipage}
	\hfill
	\begin{minipage}{0.11\linewidth}
		\centering
		\includegraphics[width=1\linewidth]{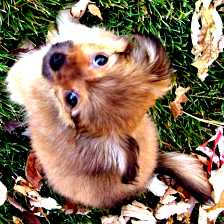}
	\end{minipage}
	\vfill
	\vspace{0.5em}
	\begin{minipage}{0.01\linewidth}
		\centering
		\footnotesize \rotatebox{90}{Oxford-Flowers}
	\end{minipage}
	\hfill
	\begin{minipage}{0.11\linewidth}
		\centering
		\includegraphics[width=1\linewidth]{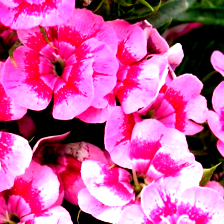}
	\end{minipage}
	\hfill
	\begin{minipage}{0.11\linewidth}
		\centering
		\includegraphics[width=1\linewidth]{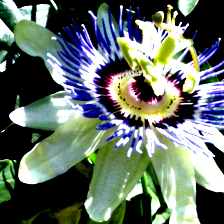}
	\end{minipage}
	\hfill
	\begin{minipage}{0.11\linewidth}
		\centering
		\includegraphics[width=1\linewidth]{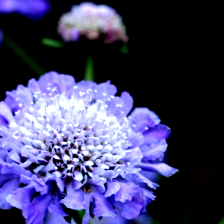}
	\end{minipage}
	\hfill
	\begin{minipage}{0.11\linewidth}
		\centering
		\includegraphics[width=1\linewidth]{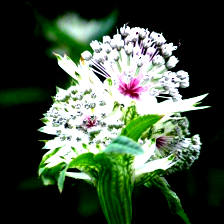}
	\end{minipage}
	\hfill
	\begin{minipage}{0.11\linewidth}
		\centering
		\includegraphics[width=1\linewidth]{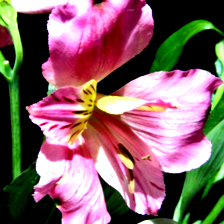}
	\end{minipage}
	\hfill
	\begin{minipage}{0.11\linewidth}
		\centering
		\includegraphics[width=1\linewidth]{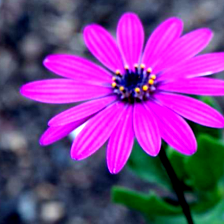}
	\end{minipage}
	\hfill
	\begin{minipage}{0.11\linewidth}
		\centering
		\includegraphics[width=1\linewidth]{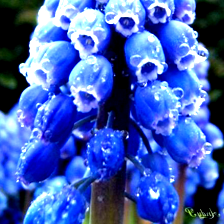}
	\end{minipage}
	\hfill
	\begin{minipage}{0.11\linewidth}
		\centering
		\includegraphics[width=1\linewidth]{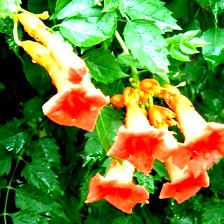}
	\end{minipage}
	\vfill
	\vspace{0.5em}
	\begin{minipage}{0.01\linewidth}
		\centering
		\footnotesize \rotatebox{90}{Food101}
	\end{minipage}
	\hfill
	\begin{minipage}{0.11\linewidth}
		\centering
		\includegraphics[width=1\linewidth]{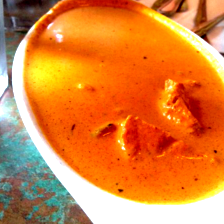}
	\end{minipage}
	\hfill
	\begin{minipage}{0.11\linewidth}
		\centering
		\includegraphics[width=1\linewidth]{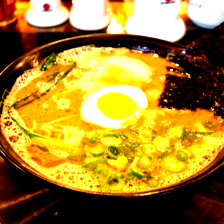}
	\end{minipage}
	\hfill
	\begin{minipage}{0.11\linewidth}
		\centering
		\includegraphics[width=1\linewidth]{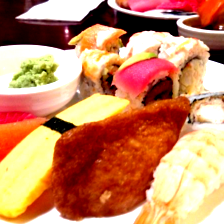}
	\end{minipage}
	\hfill
	\begin{minipage}{0.11\linewidth}
		\centering
		\includegraphics[width=1\linewidth]{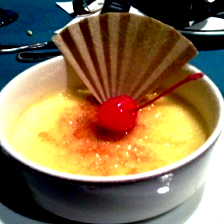}
	\end{minipage}
	\hfill
	\begin{minipage}{0.11\linewidth}
		\centering
		\includegraphics[width=1\linewidth]{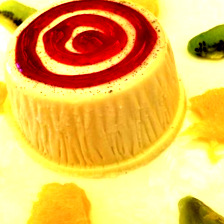}
	\end{minipage}
	\hfill
	\begin{minipage}{0.11\linewidth}
		\centering
		\includegraphics[width=1\linewidth]{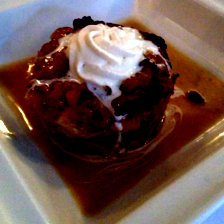}
	\end{minipage}
	\hfill
	\begin{minipage}{0.11\linewidth}
		\centering
		\includegraphics[width=1\linewidth]{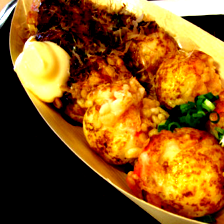}
	\end{minipage}
	\hfill
	\begin{minipage}{0.11\linewidth}
		\centering
		\includegraphics[width=1\linewidth]{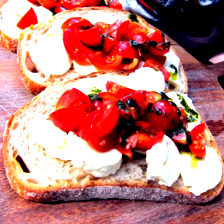}
	\end{minipage}
	\vfill
	\vspace{0.5em}
	\begin{minipage}{0.01\linewidth}
		\centering
		\footnotesize \rotatebox{90}{CIFAR100}
	\end{minipage}
	\hfill
	\begin{minipage}{0.11\linewidth}
		\centering
		\includegraphics[width=1\linewidth]{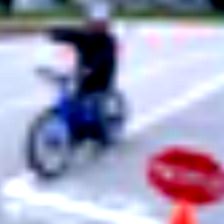}
	\end{minipage}
	\hfill
	\begin{minipage}{0.11\linewidth}
		\centering
		\includegraphics[width=1\linewidth]{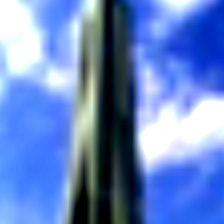}
	\end{minipage}
	\hfill
	\begin{minipage}{0.11\linewidth}
		\centering
		\includegraphics[width=1\linewidth]{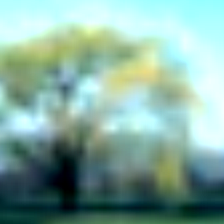}
	\end{minipage}
	\hfill
	\begin{minipage}{0.11\linewidth}
		\centering
		\includegraphics[width=1\linewidth]{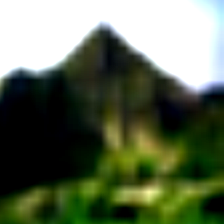}
	\end{minipage}
	\hfill
	\begin{minipage}{0.11\linewidth}
		\centering
		\includegraphics[width=1\linewidth]{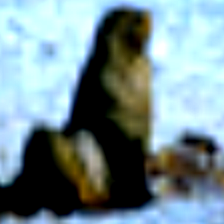}
	\end{minipage}
	\hfill
	\begin{minipage}{0.11\linewidth}
		\centering
		\includegraphics[width=1\linewidth]{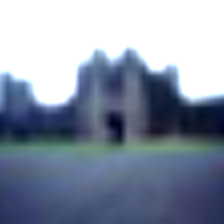}
	\end{minipage}
	\hfill
	\begin{minipage}{0.11\linewidth}
		\centering
		\includegraphics[width=1\linewidth]{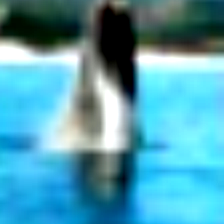}
	\end{minipage}
	\hfill
	\begin{minipage}{0.11\linewidth}
		\centering
		\includegraphics[width=1\linewidth]{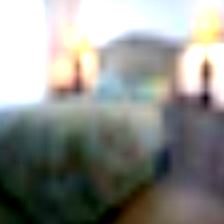}
	\end{minipage}
	\vfill
	\vspace{0.5em}
	\begin{minipage}{0.01\linewidth}
		\centering
		\footnotesize \rotatebox{90}{CIFAR10}
	\end{minipage}
	\hfill
	\begin{minipage}{0.11\linewidth}
		\centering
		\includegraphics[width=1\linewidth]{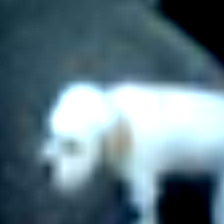}
	\end{minipage}
	\hfill
	\begin{minipage}{0.11\linewidth}
		\centering
		\includegraphics[width=1\linewidth]{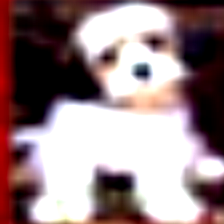}
	\end{minipage}
	\hfill
	\begin{minipage}{0.11\linewidth}
		\centering
		\includegraphics[width=1\linewidth]{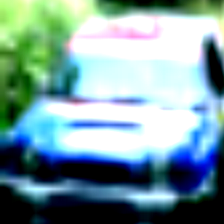}
	\end{minipage}
	\hfill
	\begin{minipage}{0.11\linewidth}
		\centering
		\includegraphics[width=1\linewidth]{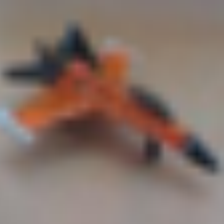}
	\end{minipage}
	\hfill
	\begin{minipage}{0.11\linewidth}
		\centering
		\includegraphics[width=1\linewidth]{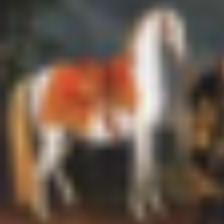}
	\end{minipage}
	\hfill
	\begin{minipage}{0.11\linewidth}
		\centering
		\includegraphics[width=1\linewidth]{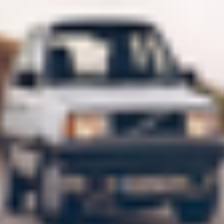}
	\end{minipage}
	\hfill
	\begin{minipage}{0.11\linewidth}
		\centering
		\includegraphics[width=1\linewidth]{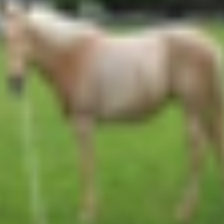}
	\end{minipage}
	\hfill
	\begin{minipage}{0.11\linewidth}
		\centering
		\includegraphics[width=1\linewidth]{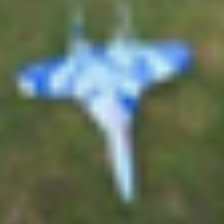}
	\end{minipage}
	\vfill
	\vspace{0.5em}
	\begin{minipage}{0.01\linewidth}
		\centering
		\footnotesize \rotatebox{90}{GTSRB}
	\end{minipage}
	\hfill
	\begin{minipage}{0.11\linewidth}
		\centering
		\includegraphics[width=1\linewidth]{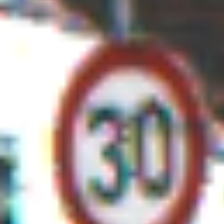}
	\end{minipage}
	\hfill
	\begin{minipage}{0.11\linewidth}
		\centering
		\includegraphics[width=1\linewidth]{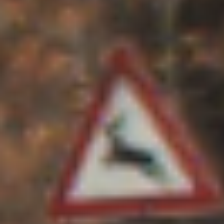}
	\end{minipage}
	\hfill
	\begin{minipage}{0.11\linewidth}
		\centering
		\includegraphics[width=1\linewidth]{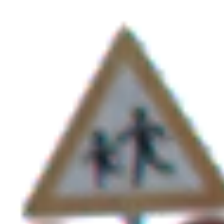}
	\end{minipage}
	\hfill
	\begin{minipage}{0.11\linewidth}
		\centering
		\includegraphics[width=1\linewidth]{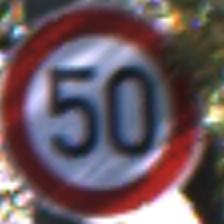}
	\end{minipage}
	\hfill
	\begin{minipage}{0.11\linewidth}
		\centering
		\includegraphics[width=1\linewidth]{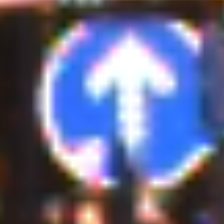}
	\end{minipage}
	\hfill
	\begin{minipage}{0.11\linewidth}
		\centering
		\includegraphics[width=1\linewidth]{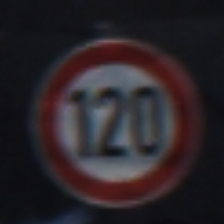}
	\end{minipage}
	\hfill
	\begin{minipage}{0.11\linewidth}
		\centering
		\includegraphics[width=1\linewidth]{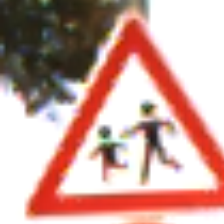}
	\end{minipage}
	\hfill
	\begin{minipage}{0.11\linewidth}
		\centering
		\includegraphics[width=1\linewidth]{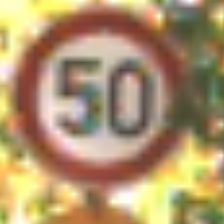}
	\end{minipage}
	\vfill
	\vspace{0.5em}
	\begin{minipage}{0.01\linewidth}
		\centering
		\footnotesize \rotatebox{90}{SVHN}
	\end{minipage}
	\hfill
	\begin{minipage}{0.11\linewidth}
		\centering
		\includegraphics[width=1\linewidth]{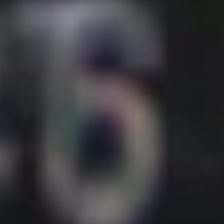}
	\end{minipage}
	\hfill
	\begin{minipage}{0.11\linewidth}
		\centering
		\includegraphics[width=1\linewidth]{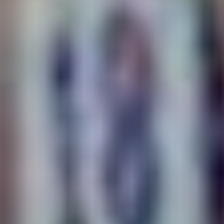}
	\end{minipage}
	\hfill
	\begin{minipage}{0.11\linewidth}
		\centering
		\includegraphics[width=1\linewidth]{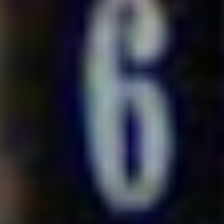}
	\end{minipage}
	\hfill
	\begin{minipage}{0.11\linewidth}
		\centering
		\includegraphics[width=1\linewidth]{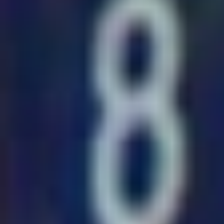}
	\end{minipage}
	\hfill
	\begin{minipage}{0.11\linewidth}
		\centering
		\includegraphics[width=1\linewidth]{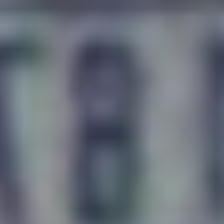}
	\end{minipage}
	\hfill
	\begin{minipage}{0.11\linewidth}
		\centering
		\includegraphics[width=1\linewidth]{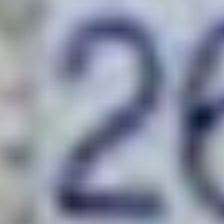}
	\end{minipage}
	\hfill
	\begin{minipage}{0.11\linewidth}
		\centering
		\includegraphics[width=1\linewidth]{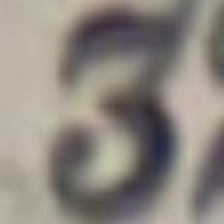}
	\end{minipage}
	\hfill
	\begin{minipage}{0.11\linewidth}
		\centering
		\includegraphics[width=1\linewidth]{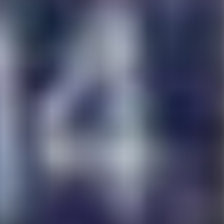}
	\end{minipage}
	\caption{Image examples for each dataset in our evaluation, where the data diversity score decreases from top to bottom.}
	\label{fig:dataset_examples}
\end{figure*}

\begin{table*}[t]
	\footnotesize
	\centering
	\setlength{\tabcolsep}{1.2mm}{
		\begin{tabular}{l|p{11mm}<{\centering}|p{9mm}<{\centering}p{9mm}<{\centering}p{9mm}<{\centering}p{9mm}<{\centering}p{9mm}<{\centering}p{9mm}<{\centering}p{11mm}<{\centering}p{11mm}<{\centering}p{9mm}<{\centering}p{9mm}<{\centering}|p{11mm}<{\centering}}
			\toprule
			& \multirow{1}{*}{Extra}
			& DTD
			& CUB200 
			& NABirds 
			& Dogs 
			& Flowers
			& Food101 
			& CIFAR100
			& CIFAR10 
			& GTSRB 
			& SVHN
			& Average
			\\
			& \multirow{1}{*}{Head}
			& \cite{dtd}
			& \cite{cub200}
			& \cite{nabirds}
			& \cite{stanford_dogs}
			& \cite{oxford_flowers}
			& \cite{food101}
			& \cite{cifar}
			& \cite{cifar}
			& \cite{gtsrb}
			& \cite{svhn}
			&
			\\
			\midrule
			Data diversity & - & 78.7 & 76.0 & 74.8 & 73.4 & 72.7 & 72.7 & 70.9 & 70.2 & 67.5 & 61.8 & -
			\\
			\midrule
			Fully-Tuning & \Checkmark & 71.3 & 78.8 & 72.8 & 89.5 & 95.1 & 83.3 & 84.0 & 97.1 & 96.8 & 90.6 & 85.9
			\\
			Linear & \Checkmark & 68.5 & 78.3 & 70.3 & 89.4 & 87.1 & 79.4 & 80.6 & 94.3 & 79.5 & 43.5 & 77.1
			\\
			Adapter \cite{adapter1,adapter2} & \Checkmark & 69.2 & 81.5 & 73.9 & 83.2 & 90.8 & 65.6 & 73.3 & 95.0 & 90.7 & 73.5 & 79.7
			\\
			VP \cite{bahng2022vp} & \Checkmark & 65.9 & 75.4 & 69.0 & 91.0 & 84.5 & 77.7 & 79.1 & 95.1 & 89.8 & 91.3 & 81.9
			\\
			VPT \cite{jia2022vpt} & \Checkmark & 67.2 & 72.1 & 65.3 & 80.5 & 88.5 & 65.2 & 72.8 & 94.4 & 88.5 & 61.8 & 75.6
			\\
			\cellcolor{gray!20}\textbf{DAM-VP (10 epochs)} & 
			\cellcolor{gray!20}\Checkmark &
			\cellcolor{gray!20}68.6 & \cellcolor{gray!20}77.0 & \cellcolor{gray!20}70.5 & \cellcolor{gray!20}93.2 & \cellcolor{gray!20}86.9 & \cellcolor{gray!20}79.6 & \cellcolor{gray!20}79.6 & \cellcolor{gray!20}95.1 & \cellcolor{gray!20}90.1 &  \cellcolor{gray!20}85.4 & \cellcolor{gray!20}82.6
			\\
			\cellcolor{gray!20}\textbf{DAM-VP (50 epochs)} & 
			\cellcolor{gray!20}\Checkmark &
			\cellcolor{gray!20}71.2 & \cellcolor{gray!20}79.7 & \cellcolor{gray!20}71.4 & \cellcolor{gray!20}93.9 & \cellcolor{gray!20}89.6 & \cellcolor{gray!20}80.1 & \cellcolor{gray!20}81.8 & \cellcolor{gray!20}95.3 & \cellcolor{gray!20}92.8 & \cellcolor{gray!20}89.3 & \cellcolor{gray!20}84.5
			\\
			\bottomrule
		\end{tabular}
	}
	\caption{Head-tuning adaption performance of different methods on MoCo-v3-B-1K, where we report image classification accuracy and all of baseline methods are trained for \textbf{100 epochs}.}
	\label{tab:moco}
\end{table*}

\begin{table*}[t]
	\footnotesize
	\centering
	\setlength{\tabcolsep}{1.2mm}{
		\begin{tabular}{l|p{11mm}<{\centering}|p{9mm}<{\centering}p{9mm}<{\centering}p{9mm}<{\centering}p{9mm}<{\centering}p{9mm}<{\centering}p{9mm}<{\centering}p{11mm}<{\centering}p{11mm}<{\centering}p{9mm}<{\centering}p{9mm}<{\centering}|p{11mm}<{\centering}}
			\toprule
			& \multirow{1}{*}{Extra}
			& DTD
			& CUB200 
			& NABirds 
			& Dogs 
			& Flowers
			& Food101 
			& CIFAR100
			& CIFAR10 
			& GTSRB 
			& SVHN
			& Average
			\\
			& \multirow{1}{*}{Head}
			& \cite{dtd}
			& \cite{cub200}
			& \cite{nabirds}
			& \cite{stanford_dogs}
			& \cite{oxford_flowers}
			& \cite{food101}
			& \cite{cifar}
			& \cite{cifar}
			& \cite{gtsrb}
			& \cite{svhn}
			&
			\\
			\midrule
			Data diversity & - & 78.7 & 76.0 & 74.8 & 73.4 & 72.7 & 72.7 & 70.9 & 70.2 & 67.5 & 61.8 & -
			\\
			\midrule
			Fully-Tuning & \Checkmark & 62.1 & 76.5 & 73.7 & 75.8 & 88.1 & 84.0 & 81.2 & 95.8 & 95.2 & 96.5 & 83.6
			\\
			Linear & \Checkmark & 64.8 & 68.1 & 58.7 & 88.5 & 81.0 & 71.8 & 71.4 & 89.9 & 79.4 & 45.3 & 71.9
			\\
			VP \cite{bahng2022vp} & \Checkmark & 63.4 & 64.3 & 56.4 & 80.7 & 78.7 & 64.2 & 62.2 & 82.1 & 84.8 & 78.1 & 71.5
			\\
			VPT \cite{jia2022vpt} & \Checkmark & 63.5 & 69.8 & 58.4 & 87.3 & 81.2 & 70.0 & 70.2 & 88.6 & 82.9 & 60.4 & 73.2
			\\
			\cellcolor{gray!20}\textbf{DAM-VP (10 epochs)} & 
			\cellcolor{gray!20}\Checkmark &
			\cellcolor{gray!20}68.4 & \cellcolor{gray!20}65.3 & \cellcolor{gray!20}57.4 & \cellcolor{gray!20}88.0 & \cellcolor{gray!20}76.1 & \cellcolor{gray!20}69.4 & \cellcolor{gray!20}71.6 & \cellcolor{gray!20}89.4 & \cellcolor{gray!20}83.7 &  \cellcolor{gray!20}75.6 & \cellcolor{gray!20}74.5
			\\
			\cellcolor{gray!20}\textbf{DAM-VP (50 epochs)} & 
			\cellcolor{gray!20}\Checkmark &
			\cellcolor{gray!20}68.5 & \cellcolor{gray!20}67.8 & \cellcolor{gray!20}58.4 & \cellcolor{gray!20}88.5 & \cellcolor{gray!20}83.7 & \cellcolor{gray!20}71.4 & \cellcolor{gray!20}72.5 & \cellcolor{gray!20}90.2 & \cellcolor{gray!20}85.6 &  \cellcolor{gray!20}78.0 & \cellcolor{gray!20}76.5
			\\
			\bottomrule
		\end{tabular}
	}
	\caption{Head-tuning adaption performance of different methods on ResNet50-1K, where we report image classification accuracy and all of baseline methods are trained for \textbf{100 epochs}.}
	\label{tab:resnet}
\end{table*}

\vspace{0.5em}
\large\noindent\textbf{E. Dataset Specification}
\vspace{0.5em}
\normalsize

We adopt total 16 datasets in experiments, in which 10 for evaluation and 6 for meta training. 
The basic information regarding these datasets is given in \Tref{tab:dataset} and image examples of evaluation datasets are showcased in \Fref{fig:dataset_examples}.

\vspace{0.5em}
\large\noindent\textbf{F. Backbone Specification}
\vspace{0.5em}
\normalsize

There are total 6 backbones are used in our experiments, shown in \Tref{tab:backbones}. 
We report the results of ViT-B-1K \cite{DosovitskiyB0WZ21}, ViT-B-22K \cite{DosovitskiyB0WZ21}, CLIP-ViT-B \cite{radford2021learning} and Swin-B-22K \cite{liu2021swin} in our manuscript and report the results of MoCo-B-1K \cite{Chen21mocov3} and  ResNet50-1K \cite{he2016deep} in this supplementary.

\vspace{0.5em}
\large\noindent\textbf{G. More Results on Other Backbones}
\vspace{0.5em}
\normalsize

\textbf{For the self-supervised pre-trained model}, we verify our DAM-VP on ViT-B/16 \cite{DosovitskiyB0WZ21} pre-trained by MoCo v3 \cite{Chen21mocov3} and show the results in \Tref{tab:moco}. 
We can find that VPT performs not good to adapt MoCo-v3 pre-trained model, whereas our DAM-VP is able to achieve comparable downstream accuracy with Full-tuning. 

\textbf{For the pre-trained convolution network}, we verify our DAM-VP on ImageNet-1k \cite{deng2009imagenet} supervised pre-trained ResNet-50 \cite{he2016deep} and show the results in \Tref{tab:resnet}. 
Note that Adapter is hard to be extended to convolution networks. 
For VPT, we follow the extending approach of its paper. 
Though obtaining lower accuracy than Full-tuning, our method still outperforms previous visual prompting methods and linear probing.

\begin{table}[htbp]
	\scriptsize
	\centering
	\setlength{\tabcolsep}{0.1mm}{
		\begin{tabular}{l|p{11mm}<{\centering}p{11mm}<{\centering}p{11mm}<{\centering}|p{11mm}<{\centering}p{11mm}<{\centering}p{11mm}<{\centering}}
			\toprule
			Backbone
			& \multicolumn{3}{c|}{ViT-B/16}
			& \multicolumn{3}{c}{ViT-L/16}
			\\
			\midrule
			VTAB-1k & Natural & Specialized & Structured & Natural & Specialized & Structured
			\\
			\midrule
			Fully-Tuning & 75.88 & 83.36 & 47.64 & 75.99 & 84.68 & 50.71
			\\
			Linear & 68.93 & 77.16 & 26.84 & 71.17 & 73.50 & 26.44
			\\
			VPT & 78.48 & 82.43 & \textbf{54.98} & 82.80 & 84.63 & 55.85
			\\
			\cellcolor{gray!20}\textbf{Ours} &  \cellcolor{gray!20}\textbf{81.29} & \cellcolor{gray!20}\textbf{83.78} & \cellcolor{gray!20}54.33 & 
			\cellcolor{gray!20}\textbf{83.53} &
			\cellcolor{gray!20}\textbf{85.24} &
			\cellcolor{gray!20}\textbf{56.35} 
			\\
			\bottomrule
		\end{tabular}
	}
	\caption{Results on VTAB benchmark (19 datasets) for ViT-B-22K and ViT-L-22K.}
	\label{tab:vtab}
\end{table}

\vspace{0.5em}
\large\noindent\textbf{H. More Results on VTAB-1k}
\vspace{0.5em}
\normalsize

VTAB-1k \cite{zhai2019large} benchmarks transfer learning methods with total 19 different task datasets, which contains three splits named ``Natural'', ``Specialized'' and ``Structured'', respectively. 
We report the comparison results in \Tref{tab:vtab}.

\vspace{0.5em}
\large\noindent\textbf{I. Hyper-Parameter Specification}
\vspace{0.5em}
\normalsize

Here we mainly specify the detailed configuration of hyper-parameters in our experiments. 
By default, we use AdamW optimizer for fully-tuning, Adapter and SGD optimizer for linear probing, VP, VPT and our DAM-VP during adaption. 
Following VPT \cite{jia2022vpt}, we adopt cosine decay scheduler and unify the warm up epochs as 10. 
The configuration about learning rate and weight decay are listed in \Tref{tab:lr_wd_0} and \ref{tab:lr_wd_1} for head-freezing/missing and head-tuning scenarios, respectively. 
During meta training, we use Reptile \cite{nichol2018reptile} as the basic solution and adopt Adam optimizer, with the unified meta learning rate (meta step size) as 0.5, the learning rate for fast update as 0.5, the unified fast update step as 4. 
The weight decay rate is set as 0 for the head-freezing/missing case and 1e-4 for the head-tuning case.

\begin{table}[t]
	\footnotesize
	\centering
	\setlength{\tabcolsep}{1mm}{
		\begin{tabular}{lp{9mm}<{\centering}p{9mm}<{\centering}p{9mm}<{\centering}p{9mm}<{\centering}p{9mm}<{\centering}}
			\toprule
			Threshold
			& 33 
			& 32
			& 31
			& 30
			& 29
			\\
			\midrule
			Flowers Acc (\%)
			& 64.3
			& 75.7 
			& 84.1 
			& 88.0 
			& 91.1
			\\
			Prompt params (M) & $0.35$ & $0.98$ & $1.82$ & $3.50$ & $4.90$ 
			\\
			\bottomrule
		\end{tabular}
	}
	\caption{\textbf{Configure clustering threshold for scaling the prompting performance. Introducing more prompts for DAM-VP benefits the accuracy when the storage is not constrained.} We test ViT-B-1K on Oxford-Flowers in the head-freezing/missing scenario. We trade-off between the accuracy and extra parameters, finally selecting 31 as the default threshold.}
	\label{tab:threshold}
\end{table}

\begin{table*}[t]
	\footnotesize
	\centering
	\setlength{\tabcolsep}{1.2mm}{
		\begin{tabular}{l|p{12mm}<{\centering}p{12mm}<{\centering}p{12mm}<{\centering}p{12mm}<{\centering}p{12mm}<{\centering}p{12mm}<{\centering}p{12mm}<{\centering}p{12mm}<{\centering}p{12mm}<{\centering}p{12mm}<{\centering}|l}
			\toprule
			\multirow{2}{*}{lr / wd}
			& DTD
			& CUB200 
			& NABirds 
			& Dogs 
			& Flowers
			& Food101 
			& CIFAR100
			& CIFAR10 
			& GTSRB 
			& SVHN
			\\
			& \cite{dtd}
			& \cite{cub200}
			& \cite{nabirds}
			& \cite{stanford_dogs}
			& \cite{oxford_flowers}
			& \cite{food101}
			& \cite{cifar}
			& \cite{cifar}
			& \cite{gtsrb}
			& \cite{svhn}
			\\
			\midrule
			Fully-Tuning & 1e-3/1e-4 & 5e-4/1e-4 & 1e-3/1e-4 & 1e-3/1e-4 & 1e-3/1e-4 & 1e-3/1e-4 & 1e-3/1e-4 & 1e-3/1e-4 & 1e-3/1e-4 & 1e-3/1e-4 & \multirow{6}{*}{\rotatebox{90}{ViT-B-1K}}
			\\
			Linear & 1e-1/0 & 5e-1/0 & 1e-3/0 & 2.5e+2/0 & 1e-1/0 & 1e-1/0 & 1e-1/0 & 1e-1/0 & 1e-1/0 & 1e-1/0 &
			\\
			Adapter \cite{adapter1,adapter2} & 5e-3/1e-4 & 1e-2/1e-1 & 5e-2/1e-2 & 5e-3/1e-2 & 1e-2/1e-2 & 5e-3/1e-4 & 5e-3/1e-4 & 1/1e-4 & 5e-1/1e-4 & 5e-1/1e-4 &
			\\
			VP \cite{bahng2022vp} & 1e+4/0 & 1e+4/0 & 1e+4/0 & 1e+4/0 & 1e+4/0 & 1e+4/0 & 1e+4/0 & 1e+4/0 & 1e+4/0 & 1e+4/0 &
			\\
			VPT \cite{jia2022vpt} & 5/1e-4 & 5e-2/1e-3 & 5/1e-4 & 5e+1/0 & 5/1e-4 & 0.25/1e-4 & 1e-2/1e-4 & 2.5/1e-2 & 5e-1/1e-4 & 2/1e-4 &
			\\
			DAM-VP & 8e+3/0 & 5e+4/0 & 1e+4/0 & 1e+4/0 & 8e+3/0 & 5e+3/0 & 5e+3/0 & 5e+3/0 & 5e+3/0 & 5e+3/0 &
			\\
			\midrule
			\midrule
			Fully-Tuning & 1e-3/1e-4 & 5e-3/1e-4 & 5e-3/1e-4 & 1e-3/1e-4 & 1e-3/1e-4 & 1e-3/1e-4 & 1e-3/1e-4 & 1e-3/1e-4 & 1e-3/1e-4 & 1e-3/1e-4 & \multirow{5}{*}{\rotatebox{90}{CLIP-ViT-B}}
			\\
			Linear & 1e-1/0 & 1e-1/0 & 1e-1/0 & 1e-1/0 & 1e-1/0 & 1e-1/0 & 1e-1/0 & 1e-1/0 & 1e-1/0 & 1e-1/0 &
			\\
			TP \cite{radford2021learning} & - & - & - & - & - & - & - & - & - & - & 
			\\
			VP \cite{bahng2022vp} & 1e+4/0 & 1e+4/0 & 1e+4/0 & 1e+4/0 & 1e+4/0 & 1e+4/0 & 1e+4/0 & 1e+4/0 & 1e+4/0 & 1e+4/0 &
			\\
			DAM-VP & 5e+4/0 & 2e+4/0 & 2e+4/0 & 1.5e+4/0 & 1e+4/0 & 5e+3/0 & 8e+3/1e-4 & 5e+3/0 & 7e+3/0 & 5e+4/0 &
			\\
			\bottomrule
		\end{tabular}
	}
	\caption{Learning rate and weight decay specification for our experiments in \textbf{head-freezing/missing} adaption.}
	\label{tab:lr_wd_0}
\end{table*}

\begin{table*}[t]
	\footnotesize
	\centering
	\setlength{\tabcolsep}{1.2mm}{
		\begin{tabular}{l|p{12mm}<{\centering}p{12mm}<{\centering}p{12mm}<{\centering}p{12mm}<{\centering}p{12mm}<{\centering}p{12mm}<{\centering}p{12mm}<{\centering}p{12mm}<{\centering}p{12mm}<{\centering}p{12mm}<{\centering}|l}
			\toprule
			\multirow{2}{*}{lr / wd}
			& DTD
			& CUB200 
			& NABirds 
			& Dogs 
			& Flowers
			& Food101 
			& CIFAR100
			& CIFAR10 
			& GTSRB 
			& SVHN
			\\
			& \cite{dtd}
			& \cite{cub200}
			& \cite{nabirds}
			& \cite{stanford_dogs}
			& \cite{oxford_flowers}
			& \cite{food101}
			& \cite{cifar}
			& \cite{cifar}
			& \cite{gtsrb}
			& \cite{svhn}
			\\
			\midrule
			Fully-Tuning & 5e-4/1e-4 & 5e-3/0 & 5e-3/0 & 5e-3/0 & 1e-3/1e-2 & 5e-4/1e-4 & 1e-3/1e-4 & 1e-3/1e-4 & 5e-4/1e-4 & 1e-3/1e-3 & \multirow{6}{*}{\rotatebox{90}{ViT-B-22K}}
			\\
			Linear & 1/0 & 5/1e-4 & 10/0 & 1e-1/1e-4 & 1e+1/1e-4 & 1e-3/0 & 1e-1/0 & 1e-2/0 & 1e-2/0 & 0.25/1e-2 &
			\\
			Adapter \cite{adapter1,adapter2} & 5e-3/1e-4 & 1e-3/1e-2 & 5e-3/1e-3 & 1e-3/1e-4 & 5e-3/1e-4 & 5e-3/1e-4 & 5e-3/1e-2 & 5e-4/1e-4 & 5e-3/1e-4 & 5e-3/1e-4 &
			\\
			VP \cite{bahng2022vp} & 4e+1/0 & 4e+1/0 & 4e+1/0 & 4e+1/0 & 4e+1/0 & 4e+1/0 & 4e+1/0 & 4e+1/0 & 4e+1/0 & 4e+1/0 &
			\\
			VPT \cite{jia2022vpt} & 5/1e-4 & 1e+1/1e-3 & 5/1e-4 & 5e+1/1e-4 & 25/1e-3 & 5/1e-3 & 5/1e-3 & 2.5/1e-2 & 1e+1/1e-4 & 2.5/0 &
			\\
			DAM-VP & 5/1e-1 & 1/1e-1 & 5/1e-2 & 1/1e-1 & 1e+1/5e-2 & 1/1e-2 & 5e-1/2e-3 & 1e-1/5e-3 & 5e+2/0 & 3e+2/0 &
			\\
			\midrule
			\midrule
			Fully-Tuning & 1e-4/1e-4 & 1e-4/1e-4 & 1e-4/1e-4 & 1e-4/1e-4 & 1e-4/1e-4 & 1e-4/1e-4 & 5e-4/1e-4 & 1e-4/1e-4 & 1e-4/1e-4 & 1e-3/1e-2 & \multirow{6}{*}{\rotatebox{90}{Swin-B-22K}}
			\\
			Linear & 2.5/1e-2 & 5e-1/0 & 5e-1/0 & 5e-1/0 & 5e-1/0 & 5e-1/0 & 1e-1/1e-2 & 5e-1/0 & 5e-1/0 & 1e-1/1e-3 &
			\\
			Adapter \cite{adapter1,adapter2} & 5e-1/1e-4 & 5e-2/1e-1 & 5e-2/1e-2 & 5e-3/1e-2 & 5e-2/1e-2 & 5e-1/1e-4 & 5e-3/1e-4 & 1/1e-4 & 5e-1/1e-4 & 5e-2/1e-4 &
			\\
			VP \cite{bahng2022vp} & 4e+1/0 & 4e+1/0 & 4e+1/0 & 4e+1/0 & 4e+1/0 & 4e+1/0 & 4e+1/0 & 4e+1/0 & 4e+1/0 & 4e+1/0 &
			\\
			VPT \cite{jia2022vpt} & 0.25/1e-2 & 5e-2/1e-3 & 5e-2/1e-3 & 5e+1/0 & 5e-2/1e-2 & 5e-3/1e-4 & 5/1e-3 & 2.5/1e-2 & 5/1e-4 & 0.25/1e-2 &
			\\
			DAM-VP & 1e-1/5e-2 & 1e-1/1e-1 & 1/1e-2 & 1e-1/1e-1 & 1/1e-4 & 1e-1/5e-2 & 5e-2/1e-2 & 5e-2/1e-2 & 5e+2/0 & 1e+1/0 &
			\\
			\midrule
			\midrule
			Fully-Tuning & 1e-3/1e-4 & 1e-3/1e-4 & 1e-3/1e-4 & 1e-3/1e-4 & 1e-3/1e-4 & 1e-3/1e-4 & 1e-3/1e-4 & 1e-3/1e-4 & 1e-3/1e-4 & 1e-3/1e-2 & \multirow{6}{*}{\rotatebox{90}{MoCo-v3-B-1K}}
			\\
			Linear & 1/0 & 1e-1/0 & 1e-1/0 & 1e-1/0 & 2.5/1e-4 & 1e-1/0 & 1e-1/0 & 1e-1/0 & 1e-1/0 & 1/0 &
			\\
			Adapter \cite{adapter1,adapter2} & 5e-3/1e-2 & 5e-2/1e-1 & 5e-2/1e-2 & 5e-3/1e-2 & 5e-3/1e-4 & 5e-1/1e-4 & 5e-3/1e-4 & 1e-2/1e-4 & 5e-1/1e-4 & 5e-3/1e-4 &
			\\
			VP \cite{bahng2022vp} & 4e+1/0 & 4e+1/0 & 4e+1/0 & 4e+1/0 & 4e+1/0 & 4e+1/0 & 4e+1/0 & 4e+1/0 & 4e+1/0 & 4e+1/0 &
			\\
			VPT \cite{jia2022vpt} & 5e+2/0 & 5e-2/1e-3 & 5e-1/1e-3 & 5/1e-4 & 1e+2/1e-4 & 1e-2/1e-4 & 1e+2/1e-4 & 1e-1/1e-3 & 2/1e-4 & 5e+1/1e-4 &
			\\
			DAM-VP & 5e-1/1e-2 & 1/5e-1 & 5/5e-2 & 1/5e-1 & 1/1e-1 & 5e-1/1e-1 & 1e-1/5e-2 & 1e-1/5e-2 & 2.5e+2/0 & 1e+1/0 &
			\\
			\midrule
			\midrule
			Fully-Tuning & 1e-3/1e-4 & 1e-3/1e-4 & 1e-3/1e-4 & 1e-3/1e-4 & 1e-3/1e-4 & 1e-3/1e-4 & 1e-3/1e-4 & 1e-3/1e-4 & 1e-3/1e-4 & 1e-3/1e-4 & \multirow{5}{*}{\rotatebox{90}{ResNet50-1K}}
			\\
			Linear & 1e-1/1e-2 & 1e-1/0 & 1e-1/0 & 1e-1/0 & 5e-2/1e-2 & 1e-1/0 & 1e-1/0 & 1e-1/0 & 1e-1/0 & 5/0 &
			\\
			VP \cite{bahng2022vp} & 1/0 & 1/0 & 1/0 & 1/0 & 1/0 & 1/0 & 1/0 & 1/0 & 1/0 & 1/0 &
			\\
			VPT \cite{jia2022vpt} & 1/1e-2 & 1e-1/1e-1 & 1/1e-2 & 1/5e-2 & 5e-1/1e-2 & 1e-2/1e-4 & 1e-1/1e-3 & 1e-1/1e-3 & 1e-1/1e-4 & 5e-1/0 &
			\\
			DAM-VP & 5e-1/5e-1 & 1e-1/1e-1 & 1/1e-2 & 1/5e-2 & 1/5e-1 & 5e-1/5e-1 & 5e-1/1e-2 & 2e-1/1e-2 & 2.5e+1/0 & 5/0 &
			\\
			\bottomrule
		\end{tabular}
	}
	\caption{Learning rate and weight decay specification for our experiments in \textbf{head-tuning} adaption.}
	\label{tab:lr_wd_1}
\end{table*}

\vspace{0.5em}
\large\noindent\textbf{J. Ablation Study on Clustering Threshold}
\vspace{0.5em}
\normalsize

We further analyse the impact of different threshold of agglomerative clustering used in our diversity-adaptive data partition. 
By default, we set the threshold as 31 for ViT-B-1K, 10 for ViT-B-22K, 20 for Swin-B-22K, 18 for MoCo-v3-B-1K and 21 for ResNet50-1K. 
Usually, the lower threshold represents the more clusters obtained by clustering. 
In \Tref{tab:threshold}, we surprisingly found that in the head-freezing/missing case, the prompting performance can be greatly boosted with the decreasing of threshold, whereas the introduced extra tunable parameters are also growing. 
It is inspiring that, especially in some cases when the storage is not a big deal, we can easily scale up the tunable parameters to get the better downstream accuracy in the head-freezing/missing scenario (almost to be closer to full-tuning performance).

\end{document}